\documentclass[runningheads]{llncs}

\usepackage{placeins}
\usepackage{eccv}

\usepackage{array}

\usepackage{eccvabbrv}

\usepackage{graphicx}
\usepackage{booktabs}
\usepackage{placeins}
  \usepackage{multirow}

\usepackage[accsupp]{axessibility}  % Improves PDF readability for those with disabilities.

\usepackage{hyperref}

\usepackage{orcidlink}

\begin{document}

\title{Evaluating Vision Foundation Models for Pixel and Object Classification in Microscopy}

\titlerunning{VFMs for Pixel and Object Classification in Microscopy}

\author{Carolin Teuber\inst{1}\orcidlink{0009-0004-3866-191X} \and
Anwai Archit\inst{1}\orcidlink{0009-0002-9533-8620} \and
Tobias Boothe\inst{2} \and
Peter Ditte \inst{2} \and
Jochen Rink \inst{2, 4} \and
Constantin Pape \inst{1, 3, 5, 6}\orcidlink{0000-0001-6562-7187}
}

\authorrunning{C.~Teuber et al.}

\institute{Georg-August-University Göttingen, Institute of Computer Science \and
Department of Tissue Dynamics and Regeneration, Max Planck Institute for Multidisciplinary Sciences, Göttingen \and
Department of Machine Intelligence in the Life Sciences, Max Planck Institute for Multidisciplinary Sciences, Göttingen \and
Georg-August-University Göttingen, Faculty of Biology and Psychology \and
CAIMed - Lower Saxony Center for AI \& Causal Methods in Medicine, Göttingen \and
Cluster of Excellence Multiscale Bioimaging (MBExC), Georg-August-University Göttingen
}

\maketitle

\begin{abstract}
Deep learning underlies most modern approaches and tools in computer vision, including biomedical imaging. However, for interactive semantic segmentation (often called \emph{pixel classification} in this context) and interactive object-level classification (\emph{object classification}), feature-based shallow learning remains widely used. This is due to the diversity of data in this domain, the lack of large pretraining datasets, and the need for computational and label efficiency. In contrast, state-of-the-art tools for many other vision tasks in microscopy — most notably cellular instance segmentation — already rely on deep learning and have recently benefited substantially from vision foundation models (VFMs), particularly SAM. Here, we investigate whether VFMs can also improve pixel and object classification compared to current approaches. To this end, we evaluate several VFMs, including general-purpose models (SAM, SAM2, SAM3, DINOv3) and domain-specific ones ($\mu$SAM, PathoSAM, KRONOS), in combination with shallow learning and attentive probing on five diverse and challenging datasets. Our results demonstrate consistent improvements over hand-crafted features and provide a clear pathway toward practical improvements. Our study also establishes a benchmark for VFMs in microscopy and informs future developments.
\end{abstract}

\section{Introduction}
\label{sec:intro}

Deep learning (DL) is the dominant paradigm in computer vision, is the state-of-the-art for most vision tasks, and has widespread real-world adoption. Until recently, most models were trained for specific tasks using supervised learning, often ImageNet-pretrained~\cite{imagenet}. Now, vision foundation models (VFMs)~\cite{sam, sam2, dinov2, dinov3} and vision–language models~\cite{clip, siglip, molmo} have enabled zero-shot prediction, in-context learning, and efficient adaptation with small adapters.

\begin{figure}[h]
  \centering
  \includegraphics[width=\linewidth]{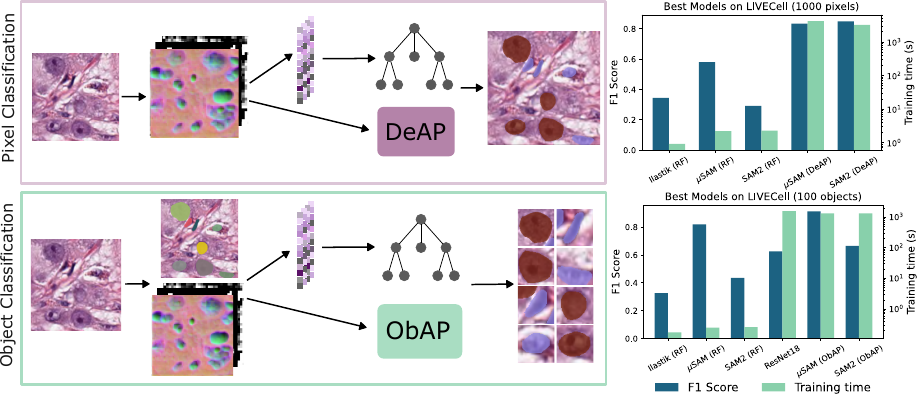}
  \caption{\textbf{Left:} \textrm{Pixel classification:} embeddings from a foundation model are used either as features for a random forest (RF) or for dense attentive probing (DeAP), both yielding pixel-level predictions. \textrm{Object classification:} embeddings are combined with instance masks to compute per-object features (for RF) or for object-guided attentive probing (ObAP), both yielding object-level predictions. \textbf{Right:} Results on LIVECell~\cite{livecell}, using 1,000 annotated pixels (top) and 100 annotated objects (bottom). DeAP and ObAP achieve the highest F1 scores, RF training is fastest.}
  \label{fig:overview}
\end{figure}

Similar to natural images, DL now underlies most analysis in biomedical imaging, driven in particular by the success of U-Net~\cite{u-net} and nnU-Net~\cite{nn-u-net}. In microscopy, cellular instance segmentation is predominantly addressed with DL, often pretrained on large and diverse datasets~\cite{cellpose, stardist}. More recently, VFMs for cell segmentation~\cite{micro-sam, cellpose, cell-sam} derived from the Segment Anything Model (SAM)~\cite{sam} have enabled zero-shot cell segmentation in many settings.

However, instance segmentation is not the only highly relevant analysis task in microscopy. So are semantic segmentation, often called \emph{pixel classification} in this context, and \emph{object classification}. Object classification refers to separating already segmented cells (or other microscopic objects) into different classes, e.g. cell types. It represents a disjoint classification stage compared to joint detection and classification in conventional \emph{object detection}. 
These tasks can be addressed with supervised DL given sufficient annotated data. However, the large diversity of tissues, cells, and sub-cellular structures across different organisms, assays, and treatments, as well as varying imaging conditions, leads to many distinct pixel and object classification tasks, for which annotated training data is lacking in many cases. This results in the continued popularity of tools using hand-crafted features and classical machine learning (ML)~\cite{ilastik, labkit, weka, qupath}. These tools enable interactive training due to the low computational and label requirements of classical ML. However, they also suffer from the limitations of classical ML, making them insufficient for complex tasks. Hence, for challenging problems, users typically need to first annotate sufficient data — often using one of the aforementioned tools — and subsequently train a DL model. This requires substantial manual effort and hinders the automation of many analysis tasks.

Consequently, a DL-based solution for interactive pixel and object classification is highly desirable. Prior work has proposed pixel classification based on features from either VFMs~\cite{convpaint, featureforest} or self-supervised learning~\cite{maester} as inputs to classical ML. However, these methods have not yet been widely adopted, and a systematic analysis of the relevant design choices is missing. Furthermore, to our knowledge, the use of VFMs for object classification has not yet been studied.

We address this research gap by systematically studying VFMs for pixel and object classification in microscopy. We analyze two key design choices:
\begin{itemize}
\item The VFM: general-purpose models (SAM~\cite{sam}, SAM2~\cite{sam2}, DINOv3~\cite{dinov3}) and domain-specific ones ($\mu$SAM~\cite{micro-sam}, PathoSAM~\cite{patho-sam}, KRONOS~\cite{shaban2025foundation}).
\item Different learning strategies, either using VFM features as input to classical ML, or training a small adapter via attentive probing.
\end{itemize}
We use Dense Attentive Probing (DeAP)~\cite{deap} for attentive probing in pixel classification and extend it to Object-Guided Attentive Probing (ObAP) for object classification.
We evaluate all approaches with respect to result quality, label efficiency, and computational efficiency. Our study therefore serves both as a benchmark for VFMs in microscopy and provides guidance for improving analysis tools. See Fig.~\ref{fig:overview} for an overview of our methodology.

\section{Related Work}

In natural images, open-vocabulary object detection and classification can be achieved with methods such as Grounding-DINO~\cite{grounding-dino} or grounded SAM~\cite{grounded-sam}, which combine the respective foundation model with CLIP~\cite{clip}. This enables the identification of arbitrary objects from a text prompt by leveraging CLIP's semantic knowledge. Similarly, SAM3~\cite{sam3} enables segmentation of specific object classes based on text prompts, relying on the text encoder of Perception Encoder~\cite{perception-encoder}. However, vision–language models trained on natural images do not contain sufficient microscopy-specific knowledge to be applicable in the use cases we study (see~\cite{micro-sam++} for an explicit study of SAM3 in microscopy). Moreover, microscopy-specific vision–language models are currently unavailable due to the lack of suitable training data.

Semantic segmentation in microscopy is typically addressed with a U-Net or related architectures~\cite{u-net, nn-u-net}, and more recently with vision transformers~\cite{vit}. Object classification can be performed with a ResNet~\cite{resnet}, or similar classification networks, applied to crops of individual cells obtained via instance segmentation. Conventional object detection is less commonly used in microscopy due to the scarcity of sufficiently annotated datasets with class labels and the availability of foundation models for cell instance segmentation, such as CellPose-SAM~\cite{cellpose-sam} and $\mu$SAM~\cite{micro-sam}, which solve many cell segmentation tasks in a zero-shot manner.

Supervised learning with a U-Net for \emph{interactive} pixel classification or a ResNet for \emph{interactive} object classification is not feasible due to the high computational and label requirements. However, interactivity is important due to the lack of sufficient annotated data for these tasks (see Introduction). Consequently, tools based on hand-crafted features and classical ML remain popular. For example, ilastik~\cite{ilastik}, LabKit~\cite{labkit}, and Weka trainable segmentation~\cite{weka} implement pixel classification using filter banks combined with a random forest. Similarly, ilastik~\cite{ilastik} implements object classification using morphological and texture features together with a random forest. Prior DL-based work has mainly focused on improving pixel classification by replacing the hand-crafted features. MAESTER~\cite{maester} trains a masked autoencoder~\cite{mae} to obtain feature representations and subsequently applies clustering in the per-pixel feature space. However, the high computational cost of training and inference makes this approach impractical for interactive use. Both ConvPaint~\cite{convpaint} and FeatureForest~\cite{featureforest} use pretrained models, including foundation models, to compute features that are fed into a shallow classifier. While they report improvements over hand-crafted features, these methods have not yet seen widespread adoption.

Despite the availability of pretrained DL and foundation models, their adoption in interactive pixel and object classification for microscopy has remained limited. We argue that a key reason is the lack of a systematic analysis of the relevant design choices. Here, we therefore perform a comprehensive study of pixel and object classification using different foundation models, learning strategies, and strong baselines, including both the best hand-crafted features and fully supervised DL.

\section{Methods}
We first summarize prior work that our methods use (Sec.~\ref{sec:prior_work}), then explain our method for pixel classification (Sec.~\ref{sec:pix}) and for object classification (Sec.~\ref{sec:object_classification}).

\subsection{Prior Work} \label{sec:prior_work}

\subsubsection{Vision Foundation Models}
refer to DL models trained on large and diverse datasets, enabling strong generalization across a wide range of downstream tasks and domains. Among the most widely used foundation models for dense prediction tasks is SAM~\cite{sam}, which supports both interactive and automatic instance segmentation. SAM consists of three main components: a vision transformer~\cite{vit} image encoder, a prompt encoder, and a mask decoder. The prompt encoder processes user inputs such as points or bounding boxes for interactive segmentation. The mask decoder combines image and prompt features to predict segmentation masks. 
SAM was trained primarily in a supervised manner with an iterative per-object segmentation objective.
The successor model, SAM2~\cite{sam2}, extends the original framework to support video and 3D data. It replaces the standard vision transformer encoder with a hierarchical vision transformer~\cite{hiera} and introduces architectural and training modifications to handle spatiotemporal data. Similar to its predecessor, SAM2 is predominantly trained using supervised learning. SAM3~\cite{sam3} unifies language and vision in a single encoder and goes beyond segmenting individually prompted objects to understanding semantic concepts, enabling it to detect, segment and track all instances of an open-vocabulary category across images and videos. 
In contrast to SAM, DINO~\cite{dino} follows a self‑supervised learning paradigm based on patch- and image-level self‑distillation. DINOv3~\cite{dinov3} significantly scales up the original DINO architecture and introduces an additional training phase aimed at preventing degradation of dense feature representations for high resolution features.

Vision foundation models generalize across many imaging modalities, including biomedical imaging, but typically perform worse than domain-specific models. To improve their applicability, SAM has been adapted to several biomedical domains, including microscopy with $\mu$SAM~\cite{micro-sam} and histopathology with PathoSAM~\cite{patho-sam}. Spatial proteomics, however, introduces additional challenges due to its high-dimensional, multi-channel data structure. KRONOS~\cite{shaban2025foundation} addresses these challenges by adapting DINOv2 with marker-specific embeddings to learn representations suited for this imaging modality.

\subsubsection{AnyUp} \label{sec:any-up}
Extracting meaningful information from VFM features requires high resolution, motivating learned upsampling methods like FeatUp~\cite{featup} or JAFAR~\cite{JAFAR}, and AnyUp~\cite{anyup}. AnyUp is trained just once and can be reused at inference time across models. Given a low resolution feature map $p \in \mathbb{R}^{h\times w \times c}$ and the corresponding high resolution image $I_{hr} \in \mathbb{R}^{H\times W}$, AnyUp produces an upsampled feature map $q = f(p, I_{hr}) \in \mathbb{R}^{H\times W \times c}$. Building on JAFAR’s pixel‑level attention, AnyUp introduces two key modifications: a feature‑agnostic convolution that maps features of arbitrary dimensionality into a shared space, and a locally restricted attention mechanism for improved efficiency. The model is trained using a combination of cosine and MSE reconstruction losses with an additional consistency regularization term.

\subsubsection{Dense Attentive Probing}\label{sec:DeAP}
\cite{deap} establishes an evaluation protocol for VFMs on dense prediction tasks. While linear and attentive probing~\cite{psomas2025attention} exist, the former is restricted to linear decision boundaries, thus, in theory, inferior to most classical ML, such as a random forest.
As shown in~\cite{deap}, neither approach is well suited for dense prediction tasks such as semantic segmentation. DeAP acts as a dense equivalent to attentive probing by training a lightweight probe on top of features from a frozen backbone. The probe operates on a fixed grid of queries, defined in the original image space, which attends to spatial locations in the features via cross-attention. To encourage spatial locality, DeAP introduces an attention mask $M(\sigma)$, parameterized by a learnable $\sigma$. The mask is modeled as a Gaussian that depends on the distance between the i-th location in the query and the j-th location in the features: 
$M_{ij} = 1 / (\sigma \sqrt{2\pi}) \exp{(-d_{ij}^2 / (2\sigma^2))}$. The outputs of each attention head are processed by a two-layer FFN, and then reshaped into a spatial tensor, which is processed by a small convolutional network to produce the dense prediction.

\subsection{Pixel Classification} \label{sec:pix}
We explore two different settings for pixel classification based on features from a foundation model: using a random forest (next paragraph) and DeAP (next subsection). We have chosen these methods to compare the quality of results, the label efficiency, and the computational efficiency of a well established classical ML algorithm with the state-of-the-art probing mechanism.

\begin{figure}[h]
  \centering
  \includegraphics[width=\linewidth]{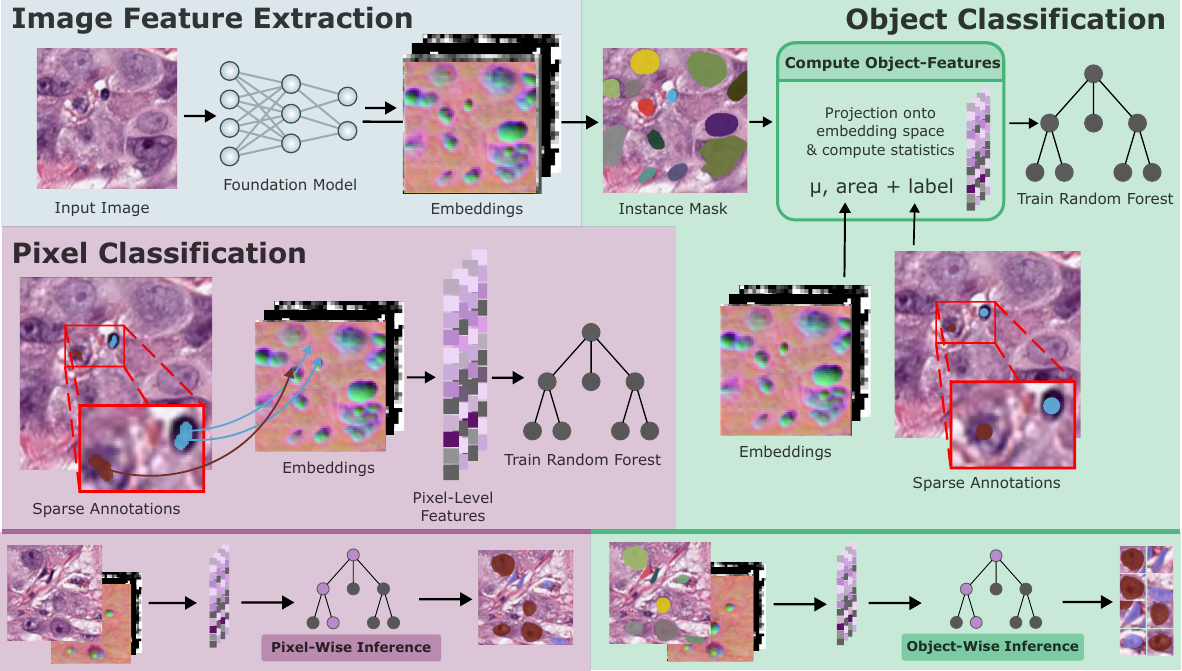}
  \caption{\textbf{Combining VFM features and RF:} For a given image, the VFM predicts dense embeddings. \textbf{Pixel classification}: sparse annotations are projected onto the embedding space, the resulting per-pixel features and labels are used to train a random forest (RF). During inference, the classifier predicts per‑pixel labels, shown as semantic segmentation. \textbf{Object classification}: instance masks are used to aggregate embeddings to obtain object‑level features. Features and object-level labels are used to train a RF. During inference, objects are classified individually with the trained RF.}
  \label{fig:random-forest}
\end{figure}

In the first setting, we build on well established tools, such as ilastik's pixel classification workflow~\cite{ilastik}, that compute pixel features via a filter bank and then train a random forest. Here, we use the image encoder of a VFM to compute features instead of a hand-crafted approach.
For training on a given image with labels, we first apply the image encoder to the image and then resize the output (embeddings) to a shape of $256\times256$ using AnyUp (see also Sec.~\ref{sec:any-up}). The labels are resized to the same shape. The (sparse) labels are projected onto the corresponding spatial location in the feature map, yielding a set of feature vectors and associated labels, as shown in Fig.~\ref{fig:random-forest}. Each vector has the same dimensionality as the embeddings. These vectors and labels are concatenated for all images in the training set and are used to train a random forest classifier. 
% In our experiments, we restrict training to a limited number of labeled pixels to mimic interactive workflows where a user would annotate the data with sparse brush-strokes. To this end, we first extract all available feature vectors from the annotated pixels and then randomly sample a fixed number of vectors for training. Sampling is performed using inverse class‑frequency weighting to ensure a balanced representation across classes in the final training set.

In our experiments, we restrict training to a limited number of labeled pixels to mimic interactive workflows in which a user annotates the data with sparse brush-strokes. We simulate these user annotations by generating scribbles as lines along an object's major axis, along with straight lines in the background, keeping all annotations at least five pixels from an object's outline. From these annotations, we extract all available feature vectors and then randomly sample a fixed number of vectors for training. Sampling is performed using inverse class-frequency weighting to ensure balanced class representation in the training set.

During inference, the trained classifier is applied independently to all pixel features, which are obtained in the same manner as before, producing per‑pixel class predictions that are aggregated into a semantic segmentation map.

\subsubsection{Pixel Classification with DeAP} \label{sec:deap_semseg}
We apply DeAP following the setup described in Sec.~\ref{sec:DeAP}. We resize all images to a shape of $1024\times1024$ before processing them with the encoder, motivated by the input size in SAM. We chose spatial dimensions $H_q \times W_q$ for the query grid, where $H_q = W_q$ are $\frac{1}{8}$ times the original input size. A CNN decoder upsamples the cross-attention outputs to the original resolution so that each query accounts for $8\times 8$ pixels in the output space.

For training, we use a weighted combination of Dice loss and cross-entropy, instead of using only the cross-entropy (as in~\cite{deap}) to account for class imbalance.
Furthermore, we explicitly support sparse pixel annotations. To this end, we incorporate a loss mask that excludes unlabeled regions from the optimization objective. 
Similar to before, we employ a pixel‑sampling strategy to control the number of supervised pixels. Given a fixed pixel budget $n_{\text{pixels}} $, we distribute the sampled pixels across the training images. If $n_{\text{pixels}} < n_{\text{images}}$, we randomly select one labeled pixel from each of $n_{\text{pixels}}$ randomly drawn images. Otherwise, multiple pixels are sampled per image. In this case, pixels are sampled in a class‑balanced manner per image by selecting an equal number of pixels per class whenever possible. If only a single pixel image is sampled, its class is chosen with a uniform probability over the number of classes.
This strategy enables effective training with partial annotations and imbalanced classes, allowing the model to learn from sparsely labeled datasets without requiring dense pixel‑wise ground truth.
In inference, the trained DeAP adapter is applied to the embeddings of an image, which are computed as in training.

\subsection{Object Classification}\label{sec:object_classification}

Our interest in object classification is partly motivated by the fact that very accurate cell instance segmentation results can be obtained with methods like $\mu$SAM~\cite{micro-sam} or CellPose-SAM~\cite{cellpose-sam} in a zero-shot setting.
Here, we therefore assume that an instance segmentation is available and focus on classifying the segmented instances. 
Following ilastik's popular object classification workflow, we first explore training a random forest classifier on object‑level features (next paragraph). 
We then introduce Object‑Guided Attentive Probing (ObAP), which extends DeAP to object classification (next subsection).

We use an existing instance segmentation to turn embeddings from a foundation model into object features. Given an input image, dense feature embeddings are first extracted using the model's encoder, following the same approach as in pixel classification (Sec.~\ref{sec:pix}). The instance masks are then used to identify individual objects in the image. For each object, the embeddings within its instance mask are aggregated channel-wise using the mean to obtain features. We add the size of the mask (area) as an additional feature (see App.~\ref{app:feature_ablation}). 
Next, the object-level class labels are matched to the features, potentially by aggregating pixel-level labels if given in that format.
This approach is repeated for all images in the training set, features and labels are sub-sampled in a class-balanced manner according to the specified label budget, and a random forest classifier is trained.
During inference, the trained classifier is applied independently to each object instance. See Fig.~\ref{fig:random-forest} for an overview of this procedure.

\subsubsection{Object-Guided Attentive Probing (ObAP)} \label{sec:obap}
\begin{figure}[h]
  \centering
  \includegraphics[width=\linewidth]{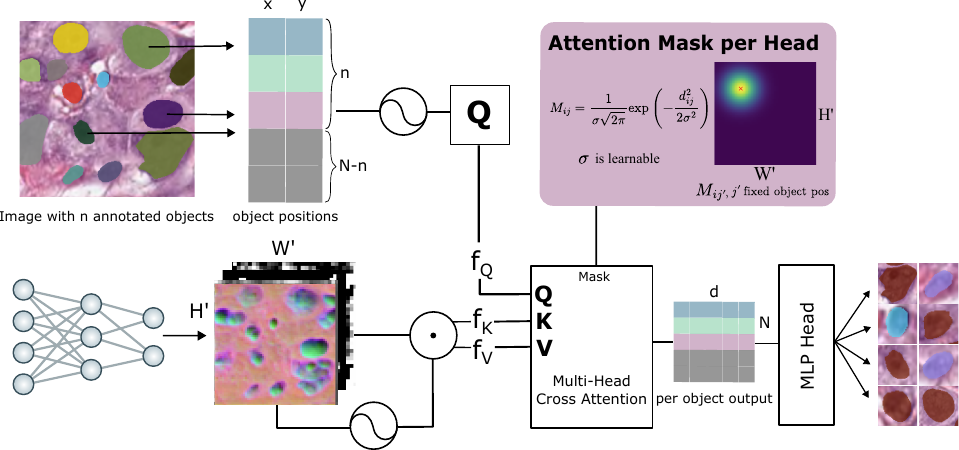}
  \caption{\textrm{\textbf{Object-guided attentive probing (ObAP):}} A VFM (frozen) extracts a feature volume from the input image. Instead of a regular grid of queries, one query is initialized per object via its center, encoded by a fixed sinusoidal positional encoding. The object queries attend to the feature volume via Gaussian‑masked cross‑attention with a learnable bandwidth parameter $\sigma$, yielding one token per object, which are then processed independently by a lightweight MLP to produce per-object class predictions.}
  \label{fig:ObAP}
\end{figure}

Building on DeAP (Sec.~ \ref{sec:DeAP}), we introduce ObAP for object classification, see Fig.~\ref{fig:ObAP} for an overview. 
Given the image $x \in \mathbb{R}^{H\times W}$, the VFM's encoder produces the feature volume $\Phi(x) \in \mathbb{R}^{H'\times W'\times C}$, as in DeAP. Instead of initializing queries on a regular grid, we use the given instances to place a query at the center of each object's mask. These positions are encoded using a fixed sinusoidal function, yielding a set of non‑learnable object queries.
To handle images containing different numbers of objects, we use a fixed‑size query tensor with a maximum number of objects. For images with fewer instances, the remaining query slots are padded and marked as invalid via a binary mask that is applied prior to cross‑attention. 
Cross‑attention then processes queries and feature volume $\Phi$. It is modulated by a Gaussian mask, as in DeAP, with $\sigma$ learned independently for each attention head. The Gaussian depends on the distance between query positions and locations in $\Phi$, to adaptively determine the spatial extent of contributing features.
The parameter $\sigma$ can thus be interpreted as the size of the embedding‑space region containing relevant information for a given object.
The cross‑attention outputs one token per object, which are processed independently with a two-layer MLP to predict the object's class. 

We train ObAP with cross entropy and use a similar sampling strategy as before to sample objects for training: Out of all instances in an image, a subset of the desired size is sampled inversely proportional to class frequencies to include underrepresented classes.
During inference, we apply the VFM's image encoder and the trained ObAP adapter. 

\section{Results}\label{sec:res}

\subsection{Datasets and Metrics}

We evaluate pixel and object classification on 5 different datasets that cover a wide range of bioimaging applications.
We have chosen this data to represent diverse use-cases from label-free microscopy, fluorescence microscopy, histopathology, and phenotypical screening.
We use fixed training, validation, and test splits for all experiments. See App.~\ref{app:data} for more details on the datasets and splits. We evaluate all results using the mean F1-score over all classes, averaged over pixels (pixel classification) or objects (object-classification). See App.~\ref{app:metrics} for details.

LIVECell~\cite{livecell} is an expert-validated phase-contrast microscopy dataset that contains over 3158 images with 1.6M individual cells from 8 different cell lines. The dataset provides both instance segmentation and cell line annotations.

We use two datasets that were acquired with the multiplexed fluorescence microscopy method CODEX. In CRC~\cite{crc}, 56 proteins, corresponding to 56  channels, were imaged in 140 tissue regions from the tumor invasive front of advanced-stage colorectal cancer. To ensure compatibility with the VFMs, we collapse the channels into a single channel by summation.
The dataset contains cell type annotations for 391,515 cells with 29 different labels, which we reduce to 9 relevant classes.  Cells are originally annotated by centroid coordinates, and we generated segmentation masks with $\mu$SAM, using the centroids as point prompts.

The HBM dataset~\cite{hbm} comprises multiplexed CODEX images of human bone marrow tissue using a 53-antibody panel. The original annotations include instance segmentations and labels for 36 unique cell types, which we aggregate into eight broader categories. The dataset contains 1,100,182 instances in total. We sum all channels into a single channel, as for CRC.

PanNuke~\cite{pannuke} is a histopathology dataset of H\&E stained images, providing annotated nuclei with instance segmentations and semantic labels across five distinct cell types. The dataset consists of 189,744 labeled nuclei in 2656 images.

The Planari dataset consists of 13,076 images of individual planarian flatworms from a phenotypic screening experiment. The images were acquired with a DSLR camera. Body plan phenotypes were experimentally altered via standard RNA interference techniques. Images are divided into two classes: wild-type and phenotypically altered worms. We segmented individual worms with SAM, using the image center as point prompt.

\subsection{Pixel Classification}
\begin{figure}[h]
  \centering
  \includegraphics[width=\linewidth]{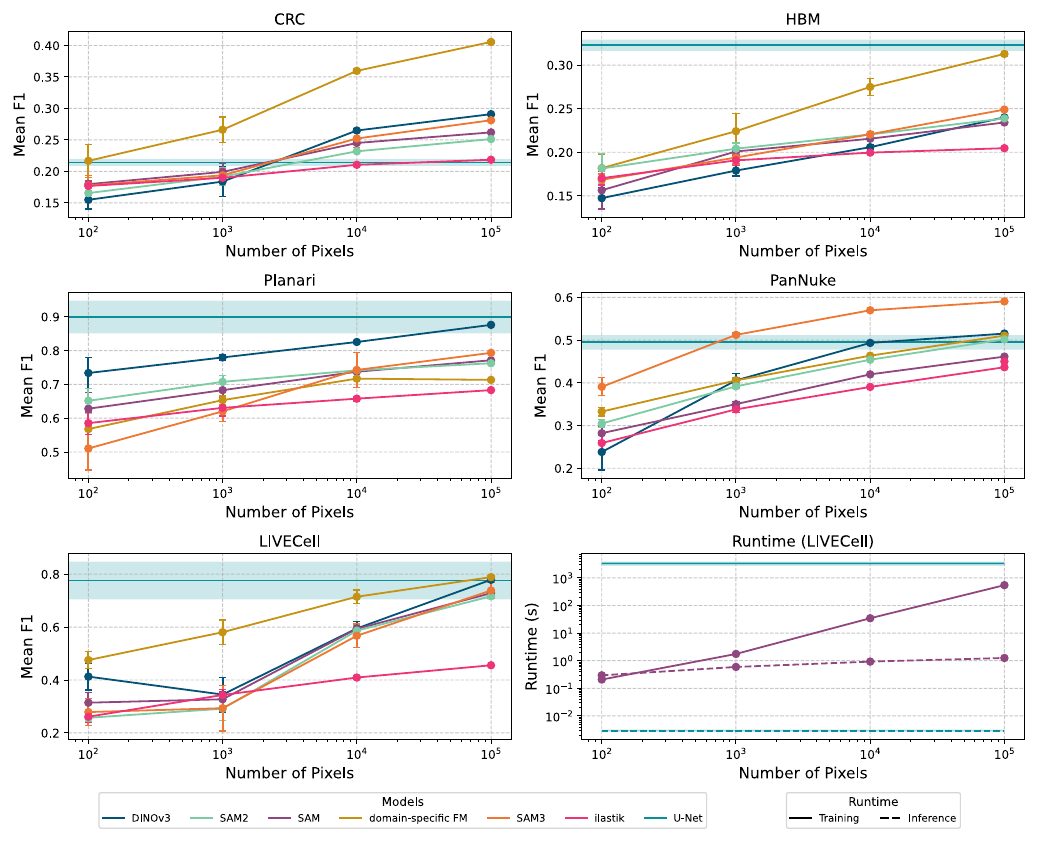}
  \caption{Pixel classification with a random forest across different training set sizes. Error bars based on 25 runs (5 folds, 5 repeats), 5 repeats for the U-Net, which is trained on all data. \textbf{Bottom right:} training (solid lines) and inference (dashed lines) times in seconds, comparing random forest (SAM features) and U-Net on LIVECell.
}
  \label{fig:pixel_classification_results}
\end{figure}

\subsubsection{Baselines:}
We compare our methods to a classical ML and a DL baseline. First, we use the hand-crafted features from ilastik pixel classification (see App.~\ref{app:classical_feats} for details) as input to a random forest. Second, a U-Net, consisting of four down- and upsampling blocks, each made of two convolutional layers followed by max pooling in the encoder and bilinear interpolation in the decoder, trained with a weighted sum of dice loss and cross entropy.

\subsubsection{With Random Forest:} \label{sec:res_pix}
We evaluate pixel classification with a random forest using general VFMs (SAM, SAM2, SAM3, DINOv3) and domain-specific VFMs (PathoSAM for PanNuke, KRONOS for CRC and HBM, otherwise $\mu$SAM) features (Fig.~\ref{fig:pixel_classification_results}, Tab.~\ref{tab:pixel_class}). See Sec.~\ref{sec:pix} for further methodology. Random forests are trained on $10^2$, $10^3$, $10^4$, and $10^5$ pixels, respectively. Each training is repeated 25 times (5 folds, 5 repeats per fold). The U-Net is trained on the full training set 5 times. Training times for the random forest correspond to fitting the training set (on CPU). Inference time is the average over images in the test set. For the U-Net, the training time is reported until convergence (on GPU), using early stopping with a patience of 10 epochs. Inference time is measured as before.

Across all datasets, VFM features clearly outperform the hand-crafted ilastik features, with the best VFM exceeding ilastik at every label budget. The domain-specific models yield the strongest results where they match the target domain: KRONOS provides large gains on CRC and HBM, and $\mu$SAM performs best on LIVECell. On Planari, where $\mu$SAM is off-domain, the general-purpose DINOv3 performs best, while SAM3 is strongest on PanNuke (note that it was trained on this data) with a strong performance of the domain-specific VFM PathoSAM for small label budgets. The fully supervised U-Net, trained on the full dataset, achieves the highest quality on Planari and HBM; on CRC, PanNuke, and LIVECell the best VFM–random forest matches or surpasses it at the higher label budgets. Its reliance on dense annotation of the full dataset nonetheless makes it impractical for the sparse, interactive setting we target. In contrast, the random forest trains within seconds on CPU, enabling interactivity. See the Appendix for qualitative results (App.~\ref{app:quali}) and an ablation of feature upsampling strategies (App.~\ref{app:anyup_results}).

%\vspace{-0.5cm}
\begin{figure}[h]
  \centering
  \includegraphics[width=\linewidth]{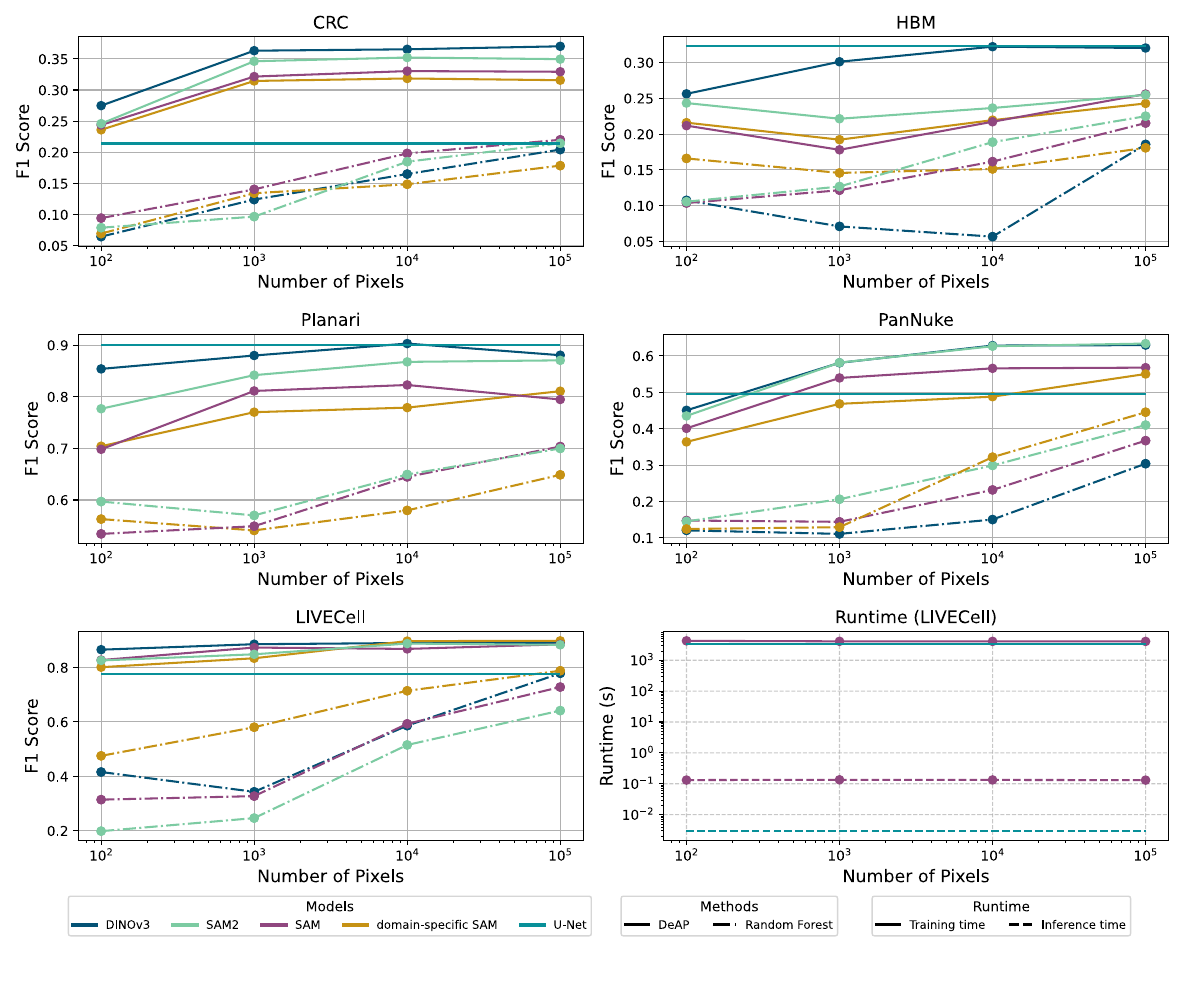}
  \caption{Pixel classification with DeAP (solid lines) compared to random forest (dash-dotted lines). \textbf{Bottom right:} training (solid lines) and inference (dashed lines) times in seconds, comparing DeAP (SAM) with U-Net on LIVECell.}
  \label{fig:deap_results}
\end{figure}

%\vspace{-1cm}
\subsubsection{With DeAP:}\label{sec:res_deap}

We further evaluate pixel classification with DeAP (Sec.~\ref{sec:DeAP}). Fig.~\ref{fig:deap_results} shows the F1 scores. 
In contrast to the random forest experiments, the DeAP results are currently limited to SAM, SAM2, DINOv3, and $\mu$SAM as the domain-specific model; SAM3 as well as KRONOS (used as the domain-specific model for CRC and HBM in the random forest experiments) are not yet included, so on CRC and HBM the domain-specific model here refers to $\mu$SAM.

Training is limited to a fixed number of pixels, as before. DeAP training is not repeated due to its long runtime. The reported training time corresponds to 10,000 iterations, inference time is recorded as before (both on GPU).

DeAP consistently outperforms Random Forest-based methods across all experiments. Notably, DeAP trained on just 100 annotated pixels achieves performance comparable to or exceeding that of Random Forest models trained on 100,000 pixels. Among all models, SAM2 and DINOv3 achieve the highest overall performance, unlike for the random forest, where the domain-specific model generally performed best. (Note though that KRONOS is missing for CRC and HBM experiments.)
On CRC, PanNuke, and LIVECell, DeAP outperforms the \emph{fully supervised} U-Net baselines. Remarkably, DeAP achieves superior performance with as few as 100 annotated pixels on LIVECell and CRC, demonstrating exceptional data efficiency and ability to generalize from limited supervision. However, DeAP's training time is only marginally faster compared to the U-Net, thus not suitable for interactivity.

\begin{figure}[h]
  \centering
  \includegraphics[width=\linewidth]{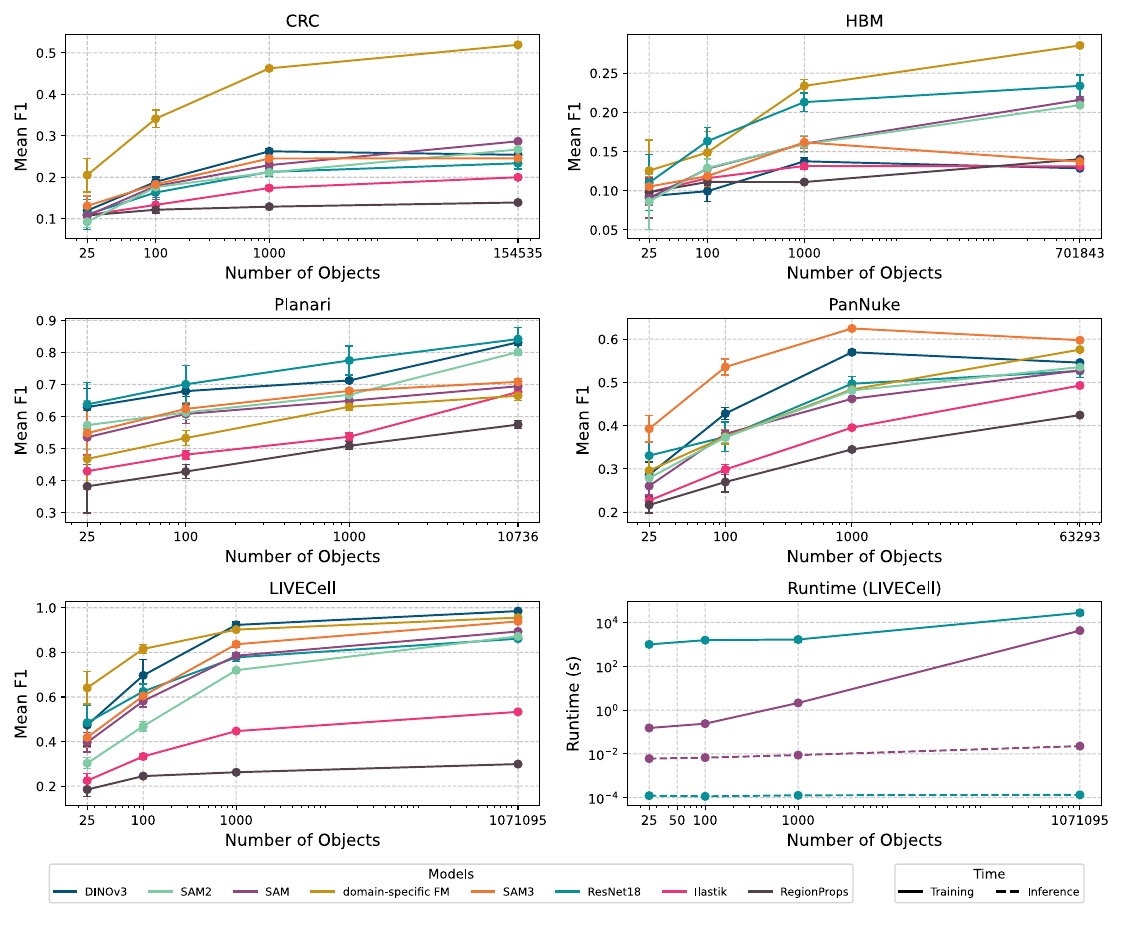}
  \caption{Object classification with a random forest, comparing VFMs, classical features and a ResNet for different training set sizes. Error bars are derived from 25 runs (five folds, five repeats). \textbf{Bottom right:} training (solid lines) and inference times (dash-dotted lines) of the ResNet and random forest (SAM features) on LIVECell.
}
\label{fig:object_classification_results}
\end{figure}

\subsection{Object Classification}

\subsubsection{Baselines:}
We compare to classical ML and DL; using the features of ilastik object classification and scikit-image~\cite{skimage} region properties (RegionProps). See App.~\ref{app:classical_feats} for a list of features.
We train a ResNet18 with ImageNet weights as DL baseline. It is trained on object-centric crops extracted around the instance masks and class labels using cross entropy. Training on crop-level enables control over the training size. Class imbalance is addressed as in the random forest training.

\subsubsection{With Random Forest:}
We evaluate object classification with a random forest based on VFM features, same models as in Sec.~\ref{sec:res_pix} (Fig.~\ref{fig:object_classification_results}, Tab. ~\ref{tab:object_class}). We use the ground-truth instance segmentation or, in the case of Planari and CRC where they are not available, instances predicted with SAM.
Classification is evaluated for training on 25, 50, 100, 1,000, and all objects. See Sec.~\ref{sec:object_classification} for further methodology.
Each experiment is repeated 25 times (5 folds, 5 repeats per fold). Runtimes for the random forest are recorded as in Sec.~\ref{sec:res_pix}. For ResNet18, training time is measured until convergence, using early stopping with a patience of 10 epochs.

Overall, VFM-derived features substantially outperform classical hand-crafted feature representations across all datasets. They are particularly effective in low-data regimes, achieving strong classification performance with only a small number of annotated objects. Domain-specific models further improve performance in selected settings, such as domain-specific SAM on LIVECell and KRONOS on the spatial proteomics datasets CRC and HBM. Nevertheless, general-purpose VFMs such as DINOv3 and SAM3 provide robust representations that generalize well across diverse biomedical imaging domains. Compared to the ResNet18 baseline, the random forest classifier operating on VFM features achieves competitive or superior performance while requiring substantially less training time, enabling more efficient and interactive object classification.

\begin{figure}[h]
  \centering
  \includegraphics[width=\linewidth]{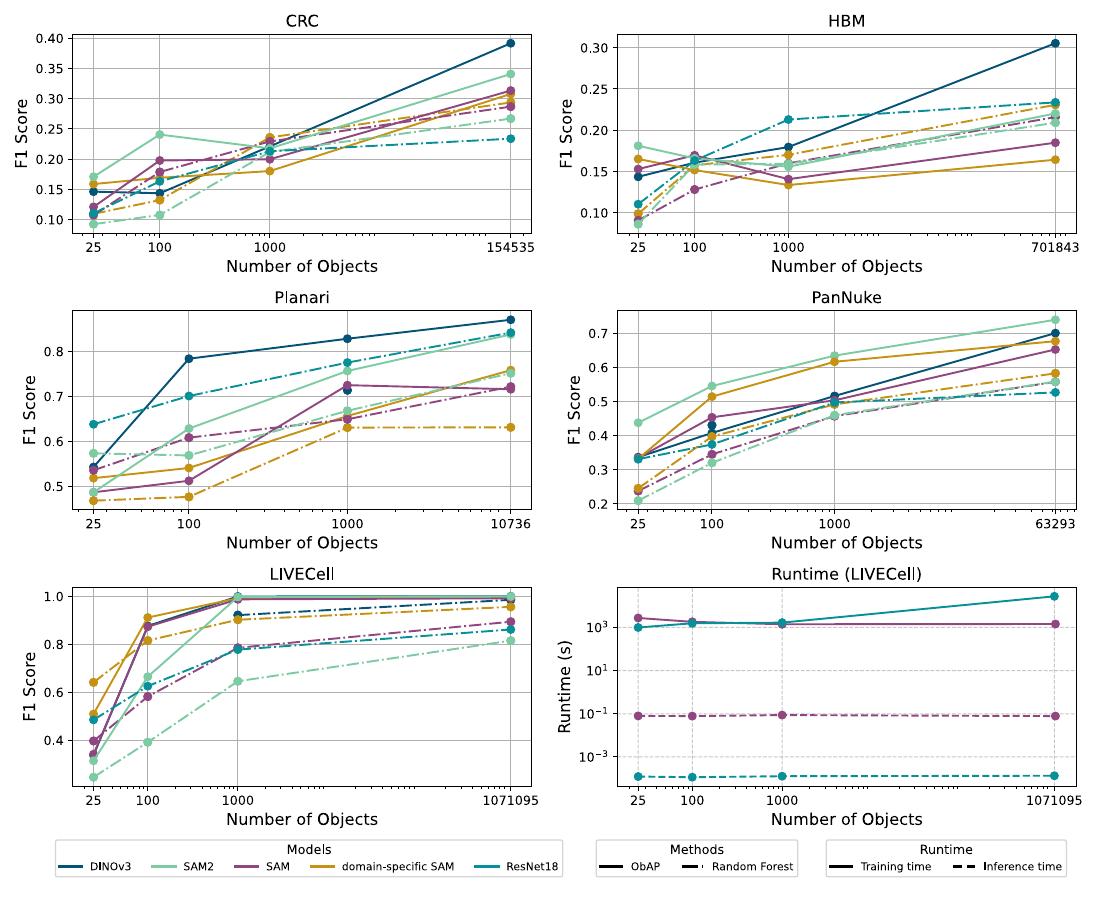}
  \caption{Performance of Object-Guided Attentive Probing for object classification (solid lines) in comparison to random forest-based object classification (dash-dotted lines). Performance is reported in F1 Score and the number of objects used for training is shown on the x-axis. The bottom right plot shows training (solid lines) and inference (dashed lines) times in seconds, comparing Object Guided Attentive Probing with SAM and the ResNet18 baseline, reported on LIVECell.}
  \label{fig:ObAP_results}
\end{figure}

\subsubsection{With ObAP:} 
We compare ObAP with the previous results (Fig.~\ref{fig:ObAP_results}), see also methodology in Sec.~\ref{sec:obap}. 

Like with DeAP, the ObAP experiments are restricted to SAM, SAM2, DINOv3, and $\mu$SAM as the domain-specific model and
do not include SAM3 or KRONOS; accordingly, the domain-specific model for CRC and HBM here is $\mu$SAM. 
We use the same set-up as before but train each ObAP model only once due to long training. Runtimes are reported as in Sec.~\ref{sec:res_deap}.

ObAP outperforms RFs in terms of result quality across most datasets and training regimes. As observed in DeAP, SAM2 and DINOv3 achieve the highest performance. 
Notably, on all datasets the best ObAP models surpass the ResNet. The training times for ObAP are comparable to those of the ResNet18 baseline. 

\section{Discussion}

We presented a comprehensive evaluation of vision foundation models (VFMs) for pixel and object classification in microscopy, comparing two learning strategies --- a random forest on VFM features and attentive probing (DeAP and ObAP) --- against classical ML and fully supervised deep learning baselines across five
diverse datasets.

Our central finding is that VFM features substantially outperform hand-crafted features in both tasks, and that attentive probing further improves over the random forest, matching or surpassing the supervised U-Net and ResNet baselines on several datasets. The two strategies favor different backbones: with the random forest, domain-specific models perform best where the domain matches
(KRONOS on CRC and HBM, $\mu$SAM on LIVECell, PathoSAM on PanNuke), whereas with attentive probing the general-purpose DINOv3 is most often the strongest backbone, with SAM2 competitive on PanNuke. Notably, DeAP and ObAP are also highly label-efficient: on several datasets a probe trained on as few as 100 annotations reaches the quality the random forest attains with orders of magnitude more labels. However, training times for DeAP and ObAP are too long to afford interactive use.

These results translate directly into practical guidance for integrating VFMs
into interactive tools such as ilastik, $\mu$SAM, or CellPose. SAM-based and DINO-based features both clearly improve over classical features, and domain-specific models are the preferred choice when combined with a random forest. Full interactivity is only possible with a random forest, attentive probing yields higher quality if longer training and higher computational demand can be afforded, potentially enabling hybrid solutions (interactive training with random forest followed by training attentive probing).

Looking ahead, the efficiency of DeAP and ObAP in low-resource settings deserves further study, as the small size of the adapters should permit efficient training, potentially even on CPU, which would enable high-quality interactive pixel and object classification. Our methodology could also be combined with further domain-specific VFMs, for example advanced histopathology models such as UNI\,v2~\cite{chen2024towards} for H\&E data, to improve results within specific domains.

Finally, the attentive probing evaluation should be extended to the full set of
backbones used with the random forest, in particular SAM3 and KRONOS. Since KRONOS natively processes the full multi-channel data via marker-specific embeddings, combining it with attentive probing could yield substantial gains on the spatial proteomics datasets.

A further direction is to extend pixel classification beyond nuclei to tissue-level classification, distinguishing larger tissue compartments and regions rather than individual cells. This is a highly relevant task in histopathology, where tissue-level region annotation underpins many diagnostic and downstream analysis workflows, and it would broaden the range of applications our approach can support.

\section*{Acknowledgements}
Anwai Archit is funded by the Deutsche Forschungsgemeinschaft (DFG, German Research Foundation) - PA 4341/2-1. Constantin Pape is supported by the German Research Foundation (Deutsche Forschungsgemeinschaft, DFG) under Germany’s Excellence Strategy - EXC 2067/1-390729940. This work is supported by the Ministry of Science and Culture of Lower Saxony through funds from the program zukunft.niedersachsen of the Volkswagen Foundation for the “CAIMed – Lower Saxony Center for Artificial Intelligence and Causal Methods in Medicine” project (grant no. ZN4257). This work is also supported by the Google Research Scholarship “Vision Foundation Models for Bioimage Segmentation”. We also gratefully acknowledge the computing time granted by the Resource Allocation Board and provided on the supercomputer Emmy at NHR@Göttingen as part of the NHR infrastructure, under the project nim00007. We thank Timo Lüddecke for his advice and support regarding the implementation of DeAP.

\newpage
\bibliographystyle{splncs04}
\bibliography{main}

\newpage

\section*{Appendix}

\subsection{Quantitative Results}

For completeness, we provide the results discussed in Sec.~\ref{sec:res} in tabular form, allowing for a more detailed, quantitative comparison than the corresponding figures.

% add tables for all experiments here
% pixel classification
\begin{table*}[t]
\centering
\caption{Semantic segmentation F1 (mean $\pm$ std over folds, where available) per number of training pixels for the foundation models (top) and the baselines (bottom). U-Net is the fully-supervised reference (budget-independent). The best model per column is marked in bold.}
\label{tab:pixel_class}
\begin{tabular}{l l cccc}
\toprule
Dataset & Model & 100 & 1{,}000 & 10{,}000 & 100{,}000 \\
\midrule
\multirow{7}{*}{LIVECell} & DINOv3 & $0.413{\scriptstyle\,\pm\,0.051}$ & $0.345{\scriptstyle\,\pm\,0.066}$ & $0.595{\scriptstyle\,\pm\,0.026}$ & $0.779{\scriptstyle\,\pm\,0.009}$ \\
 & SAM & $0.314{\scriptstyle\,\pm\,0.038}$ & $0.327{\scriptstyle\,\pm\,0.028}$ & $0.593{\scriptstyle\,\pm\,0.013}$ & $0.728{\scriptstyle\,\pm\,0.005}$ \\
 & dom.-spec. SAM & $\mathbf{0.475}{\scriptstyle\,\pm\,0.033}$ & $\mathbf{0.581}{\scriptstyle\,\pm\,0.047}$ & $\mathbf{0.715}{\scriptstyle\,\pm\,0.025}$ & $\mathbf{0.789}{\scriptstyle\,\pm\,0.007}$ \\
 & SAM2 & $0.257{\scriptstyle\,\pm\,0.014}$ & $0.292{\scriptstyle\,\pm\,0.044}$ & $0.586{\scriptstyle\,\pm\,0.020}$ & $0.716{\scriptstyle\,\pm\,0.005}$ \\
 & SAM3 & $0.279{\scriptstyle\,\pm\,0.051}$ & $0.294{\scriptstyle\,\pm\,0.086}$ & $0.567{\scriptstyle\,\pm\,0.045}$ & $0.738{\scriptstyle\,\pm\,0.030}$ \\
\cmidrule(lr){2-6}
 & Ilastik & $0.261{\scriptstyle\,\pm\,0.022}$ & $0.343{\scriptstyle\,\pm\,0.020}$ & $0.409{\scriptstyle\,\pm\,0.009}$ & $0.456{\scriptstyle\,\pm\,0.001}$ \\
 & U-Net & \multicolumn{4}{c}{$0.776{\scriptstyle\,\pm\,0.069}$} \\
\midrule
\multirow{7}{*}{Planari} & DINOv3 & $\mathbf{0.734}{\scriptstyle\,\pm\,0.046}$ & $\mathbf{0.780}{\scriptstyle\,\pm\,0.009}$ & $\mathbf{0.825}{\scriptstyle\,\pm\,0.005}$ & $\mathbf{0.876}{\scriptstyle\,\pm\,0.001}$ \\
 & SAM & $0.628{\scriptstyle\,\pm\,0.047}$ & $0.683{\scriptstyle\,\pm\,0.020}$ & $0.737{\scriptstyle\,\pm\,0.004}$ & $0.771{\scriptstyle\,\pm\,0.001}$ \\
 & dom.-spec. SAM & $0.568{\scriptstyle\,\pm\,0.017}$ & $0.653{\scriptstyle\,\pm\,0.012}$ & $0.717{\scriptstyle\,\pm\,0.006}$ & $0.713{\scriptstyle\,\pm\,0.002}$ \\
 & SAM2 & $0.651{\scriptstyle\,\pm\,0.038}$ & $0.707{\scriptstyle\,\pm\,0.019}$ & $0.742{\scriptstyle\,\pm\,0.003}$ & $0.762{\scriptstyle\,\pm\,0.001}$ \\
 & SAM3 & $0.510{\scriptstyle\,\pm\,0.063}$ & $0.620{\scriptstyle\,\pm\,0.029}$ & $0.743{\scriptstyle\,\pm\,0.051}$ & $0.793{\scriptstyle\,\pm\,0.001}$ \\
\cmidrule(lr){2-6}
 & Ilastik & $0.585{\scriptstyle\,\pm\,0.033}$ & $0.631{\scriptstyle\,\pm\,0.024}$ & $0.658{\scriptstyle\,\pm\,0.007}$ & $0.683{\scriptstyle\,\pm\,0.002}$ \\
 & U-Net & \multicolumn{4}{c}{$0.899{\scriptstyle\,\pm\,0.047}$} \\
\midrule
\multirow{7}{*}{PanNuke} & DINOv3 & $0.238{\scriptstyle\,\pm\,0.042}$ & $0.406{\scriptstyle\,\pm\,0.016}$ & $0.493{\scriptstyle\,\pm\,0.004}$ & $0.515{\scriptstyle\,\pm\,0.001}$ \\
 & SAM & $0.282{\scriptstyle\,\pm\,0.017}$ & $0.350{\scriptstyle\,\pm\,0.007}$ & $0.420{\scriptstyle\,\pm\,0.002}$ & $0.461{\scriptstyle\,\pm\,0.001}$ \\
 & dom.-spec. SAM & $0.333{\scriptstyle\,\pm\,0.010}$ & $0.406{\scriptstyle\,\pm\,0.007}$ & $0.464{\scriptstyle\,\pm\,0.002}$ & $0.510{\scriptstyle\,\pm\,0.001}$ \\
 & SAM2 & $0.305{\scriptstyle\,\pm\,0.010}$ & $0.392{\scriptstyle\,\pm\,0.006}$ & $0.454{\scriptstyle\,\pm\,0.002}$ & $0.502{\scriptstyle\,\pm\,0.001}$ \\
 & SAM3 & $\mathbf{0.391}{\scriptstyle\,\pm\,0.021}$ & $\mathbf{0.512}{\scriptstyle\,\pm\,0.004}$ & $\mathbf{0.570}{\scriptstyle\,\pm\,0.001}$ & $\mathbf{0.590}{\scriptstyle\,\pm\,0.001}$ \\
\cmidrule(lr){2-6}
 & Ilastik & $0.259{\scriptstyle\,\pm\,0.018}$ & $0.338{\scriptstyle\,\pm\,0.007}$ & $0.390{\scriptstyle\,\pm\,0.003}$ & $0.436{\scriptstyle\,\pm\,0.000}$ \\
 & U-Net & \multicolumn{4}{c}{$0.495{\scriptstyle\,\pm\,0.016}$} \\
\midrule
\multirow{8}{*}{CRC} & DINOv3 & $0.155{\scriptstyle\,\pm\,0.015}$ & $0.184{\scriptstyle\,\pm\,0.024}$ & $0.265{\scriptstyle\,\pm\,0.003}$ & $0.291{\scriptstyle\,\pm\,0.002}$ \\
 & SAM & $0.180{\scriptstyle\,\pm\,0.006}$ & $0.199{\scriptstyle\,\pm\,0.014}$ & $0.245{\scriptstyle\,\pm\,0.002}$ & $0.262{\scriptstyle\,\pm\,0.001}$ \\
 & dom.-spec. SAM & $0.184{\scriptstyle\,\pm\,0.010}$ & $0.205{\scriptstyle\,\pm\,0.011}$ & $0.254{\scriptstyle\,\pm\,0.003}$ & $0.270{\scriptstyle\,\pm\,0.001}$ \\
 & SAM2 & $0.165{\scriptstyle\,\pm\,0.009}$ & $0.192{\scriptstyle\,\pm\,0.015}$ & $0.232{\scriptstyle\,\pm\,0.002}$ & $0.251{\scriptstyle\,\pm\,0.001}$ \\
 & KRONOS & $\mathbf{0.217}{\scriptstyle\,\pm\,0.026}$ & $\mathbf{0.266}{\scriptstyle\,\pm\,0.020}$ & $\mathbf{0.360}{\scriptstyle\,\pm\,0.004}$ & $\mathbf{0.406}{\scriptstyle\,\pm\,0.003}$ \\
 & SAM3 & $0.178{\scriptstyle\,\pm\,0.016}$ & $0.194{\scriptstyle\,\pm\,0.010}$ & $0.252{\scriptstyle\,\pm\,0.015}$ & $0.281{\scriptstyle\,\pm\,0.002}$ \\
\cmidrule(lr){2-6}
 & Ilastik & $0.177{\scriptstyle\,\pm\,0.008}$ & $0.190{\scriptstyle\,\pm\,0.008}$ & $0.210{\scriptstyle\,\pm\,0.003}$ & $0.218{\scriptstyle\,\pm\,0.001}$ \\
 & U-Net & \multicolumn{4}{c}{$0.214{\scriptstyle\,\pm\,0.005}$} \\
\midrule
\multirow{8}{*}{HBM} & DINOv3 & $0.147{\scriptstyle\,\pm\,0.012}$ & $0.179{\scriptstyle\,\pm\,0.006}$ & $0.206{\scriptstyle\,\pm\,0.003}$ & $0.240{\scriptstyle\,\pm\,0.002}$ \\
 & SAM & $0.156{\scriptstyle\,\pm\,0.021}$ & $0.201{\scriptstyle\,\pm\,0.004}$ & $0.216{\scriptstyle\,\pm\,0.001}$ & $0.234{\scriptstyle\,\pm\,0.001}$ \\
 & dom.-spec. SAM & $0.182{\scriptstyle\,\pm\,0.013}$ & $0.207{\scriptstyle\,\pm\,0.006}$ & $0.227{\scriptstyle\,\pm\,0.003}$ & $0.244{\scriptstyle\,\pm\,0.001}$ \\
 & SAM2 & $0.181{\scriptstyle\,\pm\,0.017}$ & $0.204{\scriptstyle\,\pm\,0.007}$ & $0.220{\scriptstyle\,\pm\,0.001}$ & $0.239{\scriptstyle\,\pm\,0.001}$ \\
 & KRONOS & $\mathbf{0.182}{\scriptstyle\,\pm\,0.016}$ & $\mathbf{0.224}{\scriptstyle\,\pm\,0.020}$ & $\mathbf{0.275}{\scriptstyle\,\pm\,0.010}$ & $\mathbf{0.313}{\scriptstyle\,\pm\,0.002}$ \\
 & SAM3 & $0.168{\scriptstyle\,\pm\,0.007}$ & $0.194{\scriptstyle\,\pm\,0.007}$ & $0.221{\scriptstyle\,\pm\,0.003}$ & $0.249{\scriptstyle\,\pm\,0.001}$ \\
\cmidrule(lr){2-6}
 & Ilastik & $0.170{\scriptstyle\,\pm\,0.008}$ & $0.191{\scriptstyle\,\pm\,0.006}$ & $0.200{\scriptstyle\,\pm\,0.002}$ & $0.205{\scriptstyle\,\pm\,0.001}$ \\
 & U-Net & \multicolumn{4}{c}{$0.323{\scriptstyle\,\pm\,0.006}$} \\
\bottomrule
\end{tabular}
\end{table*}

% object classification
\begin{table*}[t]
\centering
\caption{Object classification F1 (mean $\pm$ std over folds, where available) per number of training objects for foundation models (top) and baselines (bottom). The best model per column is marked in bold.}
\label{tab:object_class}
\begin{tabular}{l l cccc}
\toprule
Dataset & Model & 25 & 100 & 1{,}000 & all \\
\midrule
\multirow{8}{*}{LIVECell} & DINOv3 & $0.474{\scriptstyle\,\pm\,0.094}$ & $0.697{\scriptstyle\,\pm\,0.072}$ & $\mathbf{0.923}{\scriptstyle\,\pm\,0.013}$ & $\mathbf{0.985}{\scriptstyle\,\pm\,0.000}$ \\
 & SAM & $0.396{\scriptstyle\,\pm\,0.043}$ & $0.582{\scriptstyle\,\pm\,0.026}$ & $0.785{\scriptstyle\,\pm\,0.004}$ & $0.894{\scriptstyle\,\pm\,0.000}$ \\
 & dom.-spec. SAM & $\mathbf{0.641}{\scriptstyle\,\pm\,0.072}$ & $\mathbf{0.815}{\scriptstyle\,\pm\,0.020}$ & $0.902{\scriptstyle\,\pm\,0.005}$ & $0.956{\scriptstyle\,\pm\,0.000}$ \\
 & SAM2 & $0.303{\scriptstyle\,\pm\,0.025}$ & $0.469{\scriptstyle\,\pm\,0.021}$ & $0.720{\scriptstyle\,\pm\,0.009}$ & $0.872{\scriptstyle\,\pm\,0.000}$ \\
 & SAM3 & $0.419{\scriptstyle\,\pm\,0.044}$ & $0.605{\scriptstyle\,\pm\,0.028}$ & $0.836{\scriptstyle\,\pm\,0.013}$ & $0.939{\scriptstyle\,\pm\,0.000}$ \\
\cmidrule(lr){2-6}
 & ResNet18 & $0.485{\scriptstyle\,\pm\,0.077}$ & $0.625{\scriptstyle\,\pm\,0.033}$ & $0.778{\scriptstyle\,\pm\,0.020}$ & $0.861{\scriptstyle\,\pm\,0.009}$ \\
 & Ilastik & $0.225{\scriptstyle\,\pm\,0.032}$ & $0.333{\scriptstyle\,\pm\,0.014}$ & $0.447{\scriptstyle\,\pm\,0.004}$ & $0.533{\scriptstyle\,\pm\,0.000}$ \\
 & RegionProps & $0.185{\scriptstyle\,\pm\,0.031}$ & $0.245{\scriptstyle\,\pm\,0.005}$ & $0.263{\scriptstyle\,\pm\,0.004}$ & $0.299{\scriptstyle\,\pm\,0.000}$ \\
\midrule
\multirow{8}{*}{Planari} & DINOv3 & $0.630{\scriptstyle\,\pm\,0.057}$ & $0.679{\scriptstyle\,\pm\,0.018}$ & $0.713{\scriptstyle\,\pm\,0.006}$ & $0.832{\scriptstyle\,\pm\,0.010}$ \\
 & SAM & $0.535{\scriptstyle\,\pm\,0.084}$ & $0.608{\scriptstyle\,\pm\,0.029}$ & $0.649{\scriptstyle\,\pm\,0.009}$ & $0.695{\scriptstyle\,\pm\,0.008}$ \\
 & dom.-spec. SAM & $0.468{\scriptstyle\,\pm\,0.094}$ & $0.533{\scriptstyle\,\pm\,0.024}$ & $0.630{\scriptstyle\,\pm\,0.012}$ & $0.665{\scriptstyle\,\pm\,0.014}$ \\
 & SAM2 & $0.573{\scriptstyle\,\pm\,0.076}$ & $0.613{\scriptstyle\,\pm\,0.023}$ & $0.668{\scriptstyle\,\pm\,0.013}$ & $0.801{\scriptstyle\,\pm\,0.009}$ \\
 & SAM3 & $0.548{\scriptstyle\,\pm\,0.078}$ & $0.624{\scriptstyle\,\pm\,0.019}$ & $0.680{\scriptstyle\,\pm\,0.007}$ & $0.708{\scriptstyle\,\pm\,0.011}$ \\
\cmidrule(lr){2-6}
 & ResNet18 & $\mathbf{0.638}{\scriptstyle\,\pm\,0.068}$ & $\mathbf{0.701}{\scriptstyle\,\pm\,0.058}$ & $\mathbf{0.775}{\scriptstyle\,\pm\,0.046}$ & $\mathbf{0.842}{\scriptstyle\,\pm\,0.036}$ \\
 & Ilastik & $0.430{\scriptstyle\,\pm\,0.050}$ & $0.481{\scriptstyle\,\pm\,0.013}$ & $0.537{\scriptstyle\,\pm\,0.012}$ & $0.677{\scriptstyle\,\pm\,0.013}$ \\
 & RegionProps & $0.382{\scriptstyle\,\pm\,0.083}$ & $0.428{\scriptstyle\,\pm\,0.021}$ & $0.509{\scriptstyle\,\pm\,0.013}$ & $0.575{\scriptstyle\,\pm\,0.013}$ \\
\midrule
\multirow{8}{*}{PanNuke} & DINOv3 & $0.287{\scriptstyle\,\pm\,0.030}$ & $0.428{\scriptstyle\,\pm\,0.014}$ & $0.570{\scriptstyle\,\pm\,0.005}$ & $0.546{\scriptstyle\,\pm\,0.001}$ \\
 & SAM & $0.261{\scriptstyle\,\pm\,0.022}$ & $0.381{\scriptstyle\,\pm\,0.009}$ & $0.462{\scriptstyle\,\pm\,0.004}$ & $0.528{\scriptstyle\,\pm\,0.002}$ \\
 & dom.-spec. SAM & $0.296{\scriptstyle\,\pm\,0.037}$ & $0.375{\scriptstyle\,\pm\,0.017}$ & $0.484{\scriptstyle\,\pm\,0.006}$ & $0.576{\scriptstyle\,\pm\,0.002}$ \\
 & SAM2 & $0.279{\scriptstyle\,\pm\,0.017}$ & $0.373{\scriptstyle\,\pm\,0.012}$ & $0.481{\scriptstyle\,\pm\,0.005}$ & $0.535{\scriptstyle\,\pm\,0.002}$ \\
 & SAM3 & $\mathbf{0.393}{\scriptstyle\,\pm\,0.031}$ & $\mathbf{0.536}{\scriptstyle\,\pm\,0.018}$ & $\mathbf{0.625}{\scriptstyle\,\pm\,0.002}$ & $\mathbf{0.598}{\scriptstyle\,\pm\,0.001}$ \\
\cmidrule(lr){2-6}
 & ResNet18 & $0.331{\scriptstyle\,\pm\,0.032}$ & $0.374{\scriptstyle\,\pm\,0.034}$ & $0.497{\scriptstyle\,\pm\,0.017}$ & $0.527{\scriptstyle\,\pm\,0.014}$ \\
 & Ilastik & $0.227{\scriptstyle\,\pm\,0.013}$ & $0.299{\scriptstyle\,\pm\,0.011}$ & $0.395{\scriptstyle\,\pm\,0.004}$ & $0.493{\scriptstyle\,\pm\,0.001}$ \\
 & RegionProps & $0.217{\scriptstyle\,\pm\,0.018}$ & $0.270{\scriptstyle\,\pm\,0.023}$ & $0.345{\scriptstyle\,\pm\,0.004}$ & $0.424{\scriptstyle\,\pm\,0.001}$ \\
\midrule
\multirow{9}{*}{CRC} & DINOv3 & $0.120{\scriptstyle\,\pm\,0.014}$ & $0.189{\scriptstyle\,\pm\,0.012}$ & $0.263{\scriptstyle\,\pm\,0.006}$ & $0.254{\scriptstyle\,\pm\,0.002}$ \\
 & SAM & $0.107{\scriptstyle\,\pm\,0.014}$ & $0.179{\scriptstyle\,\pm\,0.016}$ & $0.229{\scriptstyle\,\pm\,0.007}$ & $0.287{\scriptstyle\,\pm\,0.002}$ \\
 & dom.-spec. SAM & $0.109{\scriptstyle\,\pm\,0.018}$ & $0.175{\scriptstyle\,\pm\,0.008}$ & $0.236{\scriptstyle\,\pm\,0.007}$ & $0.294{\scriptstyle\,\pm\,0.003}$ \\
 & SAM2 & $0.093{\scriptstyle\,\pm\,0.014}$ & $0.176{\scriptstyle\,\pm\,0.010}$ & $0.212{\scriptstyle\,\pm\,0.007}$ & $0.267{\scriptstyle\,\pm\,0.002}$ \\
 & KRONOS & $\mathbf{0.205}{\scriptstyle\,\pm\,0.040}$ & $\mathbf{0.341}{\scriptstyle\,\pm\,0.021}$ & $\mathbf{0.463}{\scriptstyle\,\pm\,0.003}$ & $\mathbf{0.519}{\scriptstyle\,\pm\,0.002}$ \\
 & SAM3 & $0.131{\scriptstyle\,\pm\,0.024}$ & $0.183{\scriptstyle\,\pm\,0.013}$ & $0.245{\scriptstyle\,\pm\,0.006}$ & $0.246{\scriptstyle\,\pm\,0.002}$ \\
\cmidrule(lr){2-6}
 & ResNet18 & $0.110{\scriptstyle\,\pm\,0.036}$ & $0.163{\scriptstyle\,\pm\,0.018}$ & $0.213{\scriptstyle\,\pm\,0.012}$ & $0.234{\scriptstyle\,\pm\,0.014}$ \\
 & Ilastik & $0.108{\scriptstyle\,\pm\,0.006}$ & $0.133{\scriptstyle\,\pm\,0.018}$ & $0.174{\scriptstyle\,\pm\,0.007}$ & $0.200{\scriptstyle\,\pm\,0.002}$ \\
 & RegionProps & $0.108{\scriptstyle\,\pm\,0.018}$ & $0.122{\scriptstyle\,\pm\,0.009}$ & $0.129{\scriptstyle\,\pm\,0.003}$ & $0.139{\scriptstyle\,\pm\,0.002}$ \\
\midrule
\multirow{9}{*}{HBM} & DINOv3 & $0.093{\scriptstyle\,\pm\,0.010}$ & $0.099{\scriptstyle\,\pm\,0.013}$ & $0.138{\scriptstyle\,\pm\,0.005}$ & $0.129{\scriptstyle\,\pm\,0.000}$ \\
 & SAM & $0.091{\scriptstyle\,\pm\,0.026}$ & $0.128{\scriptstyle\,\pm\,0.012}$ & $0.160{\scriptstyle\,\pm\,0.009}$ & $0.216{\scriptstyle\,\pm\,0.001}$ \\
 & dom.-spec. SAM & $0.099{\scriptstyle\,\pm\,0.027}$ & $0.128{\scriptstyle\,\pm\,0.015}$ & $0.170{\scriptstyle\,\pm\,0.013}$ & $0.231{\scriptstyle\,\pm\,0.001}$ \\
 & SAM2 & $0.086{\scriptstyle\,\pm\,0.035}$ & $0.129{\scriptstyle\,\pm\,0.012}$ & $0.159{\scriptstyle\,\pm\,0.011}$ & $0.209{\scriptstyle\,\pm\,0.000}$ \\
 & KRONOS & $\mathbf{0.126}{\scriptstyle\,\pm\,0.039}$ & $0.149{\scriptstyle\,\pm\,0.027}$ & $\mathbf{0.234}{\scriptstyle\,\pm\,0.008}$ & $\mathbf{0.285}{\scriptstyle\,\pm\,0.002}$ \\
 & SAM3 & $0.105{\scriptstyle\,\pm\,0.013}$ & $0.119{\scriptstyle\,\pm\,0.014}$ & $0.162{\scriptstyle\,\pm\,0.007}$ & $0.137{\scriptstyle\,\pm\,0.001}$ \\
\cmidrule(lr){2-6}
 & ResNet18 & $0.110{\scriptstyle\,\pm\,0.036}$ & $\mathbf{0.163}{\scriptstyle\,\pm\,0.018}$ & $0.213{\scriptstyle\,\pm\,0.012}$ & $0.234{\scriptstyle\,\pm\,0.014}$ \\
 & Ilastik & $0.098{\scriptstyle\,\pm\,0.016}$ & $0.116{\scriptstyle\,\pm\,0.005}$ & $0.132{\scriptstyle\,\pm\,0.004}$ & $0.131{\scriptstyle\,\pm\,0.001}$ \\
 & RegionProps & $0.098{\scriptstyle\,\pm\,0.015}$ & $0.111{\scriptstyle\,\pm\,0.004}$ & $0.111{\scriptstyle\,\pm\,0.002}$ & $0.140{\scriptstyle\,\pm\,0.000}$ \\
\bottomrule
\end{tabular}
\end{table*}

%DeAP
\begin{table*}[t]
\centering
\caption{Semantic segmentation F1 (mean over folds) for DeAP per number of training pixels. The best backbone per column is marked in bold. }
\label{tab:deap_pixels}
\begin{tabular}{l l cccc}
\toprule
Dataset & Model & 100 & 1{,}000 & 10{,}000 & 100{,}000 \\
\midrule
\multirow{4}{*}{LIVECell} & DINOv3 & \textbf{0.866} & \textbf{0.886} & 0.890 & 0.890 \\
 & SAM & 0.827 & 0.873 & 0.868 & 0.885 \\
 & dom.-spec. SAM & 0.801 & 0.834 & \textbf{0.897} & \textbf{0.898} \\
 & SAM2 & 0.826 & 0.849 & 0.888 & 0.883 \\
\midrule
\multirow{4}{*}{Planari} & DINOv3 & \textbf{0.854} & \textbf{0.880} & \textbf{0.903} & \textbf{0.880} \\
 & SAM & 0.698 & 0.811 & 0.823 & 0.795 \\
 & dom.-spec. SAM & 0.704 & 0.770 & 0.779 & 0.811 \\
 & SAM2 & 0.777 & 0.842 & 0.867 & 0.870 \\
\midrule
\multirow{4}{*}{PanNuke} & DINOv3 & \textbf{0.450} & 0.581 & \textbf{0.628} & 0.630 \\
 & SAM & 0.401 & 0.539 & 0.566 & 0.568 \\
 & dom.-spec. SAM & 0.364 & 0.468 & 0.488 & 0.550 \\
 & SAM2 & 0.435 & \textbf{0.582} & 0.627 & \textbf{0.634} \\
\midrule
\multirow{4}{*}{CRC} & DINOv3 & \textbf{0.275} & \textbf{0.363} & \textbf{0.365} & \textbf{0.370} \\
 & SAM & 0.244 & 0.322 & 0.330 & 0.329 \\
 & dom.-spec. SAM & 0.236 & 0.315 & 0.319 & 0.316 \\
 & SAM2 & 0.246 & 0.346 & 0.352 & 0.350 \\
\midrule
\multirow{4}{*}{HBM} & DINOv3 & \textbf{0.256} & \textbf{0.301} & \textbf{0.322} & \textbf{0.320} \\
 & SAM & 0.212 & 0.178 & 0.217 & 0.256 \\
 & dom.-spec. SAM & 0.216 & 0.192 & 0.219 & 0.243 \\
 & SAM2 & 0.243 & 0.221 & 0.236 & 0.255 \\
\bottomrule
\end{tabular}
\end{table*}

% deap objects
\begin{table*}[t]
\centering
\caption{Object classification F1 (mean over folds) for ObAP per number of training objects. The best backbone per column is marked in bold. }
\label{tab:deap_objects}
\begin{tabular}{l l cccc}
\toprule
Dataset & Model & 25 & 100 & 1{,}000 & all \\
\midrule
\multirow{4}{*}{LIVECell} & DINOv3 & 0.337 & 0.876 & \textbf{0.999} & \textbf{1.000} \\
 & SAM & 0.341 & 0.873 & 0.987 & 0.992 \\
 & dom.-spec. SAM & \textbf{0.509} & \textbf{0.911} & 0.992 & 0.994 \\
 & SAM2 & 0.314 & 0.664 & 0.997 & 0.999 \\
\midrule
\multirow{4}{*}{Planari} & DINOv3 & \textbf{0.543} & \textbf{0.784} & \textbf{0.828} & \textbf{0.870} \\
 & SAM & 0.486 & 0.512 & 0.725 & 0.716 \\
 & dom.-spec. SAM & 0.518 & 0.541 & 0.656 & 0.758 \\
 & SAM2 & 0.486 & 0.628 & 0.756 & 0.838 \\
\midrule
\multirow{4}{*}{PanNuke} & DINOv3 & 0.337 & 0.408 & 0.516 & 0.700 \\
 & SAM & 0.334 & 0.454 & 0.503 & 0.652 \\
 & dom.-spec. SAM & 0.332 & 0.514 & 0.616 & 0.677 \\
 & SAM2 & \textbf{0.438} & \textbf{0.545} & \textbf{0.634} & \textbf{0.739} \\
\midrule
\multirow{4}{*}{CRC} & DINOv3 & 0.146 & 0.144 & \textbf{0.221} & \textbf{0.392} \\
 & SAM & 0.121 & 0.198 & 0.200 & 0.313 \\
 & dom.-spec. SAM & 0.159 & 0.169 & 0.180 & 0.308 \\
 & SAM2 & \textbf{0.171} & \textbf{0.241} & 0.218 & 0.341 \\
\midrule
\multirow{4}{*}{HBM} & DINOv3 & 0.144 & 0.161 & \textbf{0.180} & \textbf{0.305} \\
 & SAM & 0.153 & \textbf{0.170} & 0.141 & 0.185 \\
 & dom.-spec. SAM & 0.165 & 0.152 & 0.134 & 0.164 \\
 & SAM2 & \textbf{0.181} & 0.166 & 0.156 & 0.220 \\
\bottomrule
\end{tabular}
\end{table*}

\subsection{Qualitative Results}\label{app:quali}
\begin{figure}[h!]
  \centering
  \includegraphics[width=\linewidth]{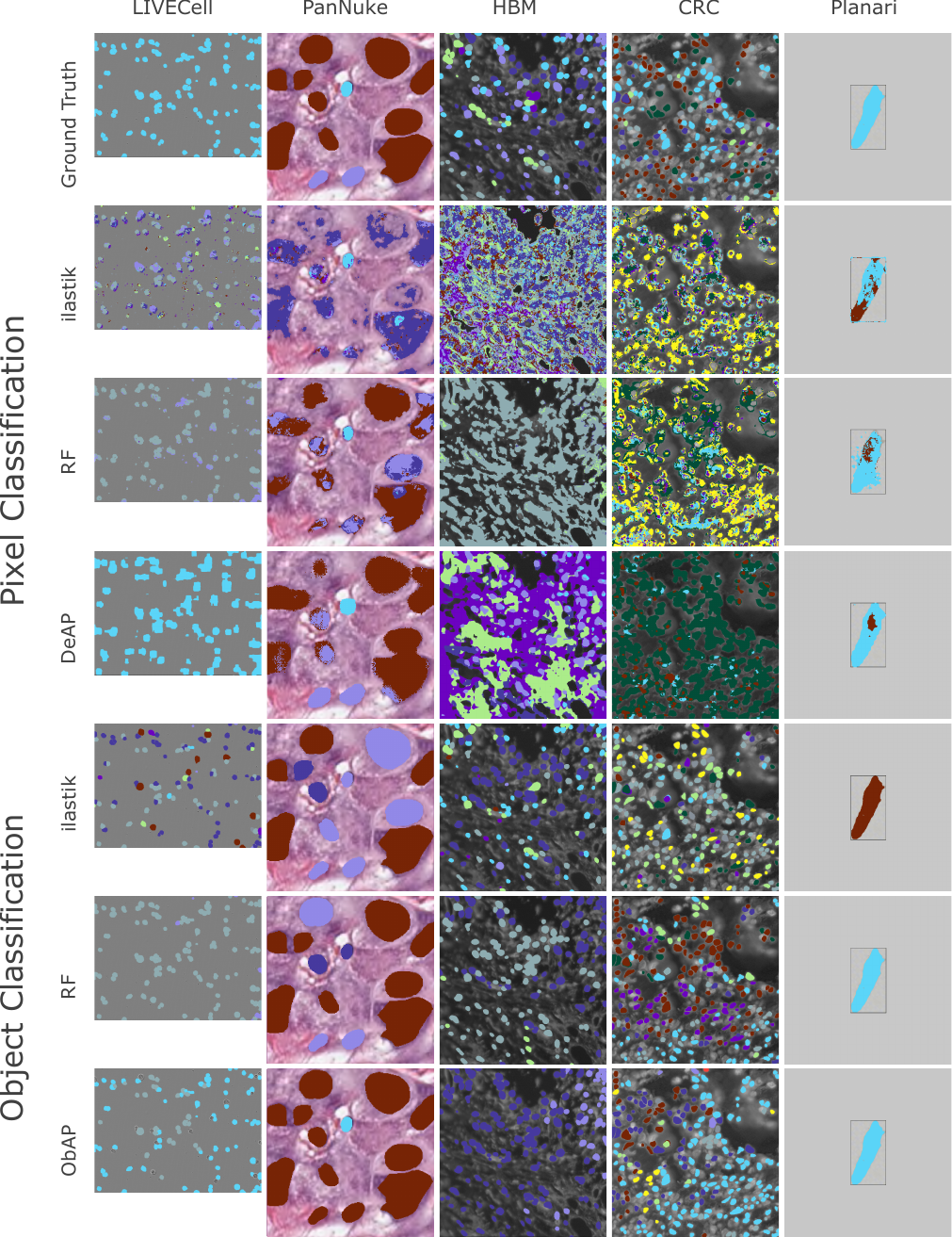}
  \caption{Qualitative results for pixel‑ and object classification across datasets using domain-specific SAM features. For object classification, models were trained using 100 annotated objects; pixel classification models were trained with 1000 annotated pixels. For object classification, the semantic segmentations are obtained by projecting the predicted object classes onto instances (ground‑truth instance masks where available; $\mu$SAM / SAM‑based for CRC and Planaria).}
  \label{fig:quali}
\end{figure}

 Fig. \ref{fig:quali} presents qualitative examples of the predictions obtained with pixel‑level and object‑level classification models across the considered datasets. For each experiment, domain-specific SAM features were used (PathoSAM for PanNuke and $\mu$SAM elsewhere) and the classical features from ilastik are added as a baseline.
 For object classification, semantic segmentation visualizations are generated by assigning the predicted class labels to instance segmentations, using ground‑truth instance masks when available and $\mu$SAM-based instance segmentations for CRC and Planaria.
 
 % The examples illustrate typical prediction behavior and provide an intuitive comparison between pixel‑ and object‑based approaches. Note that for the pixel classification background is segmented as a foreground class in some cases (LIVECell and HBM). This is likely due to densely growing cells and due to the fact that we do not oversample the background class with respect to the different foreground classes. Further, note that interactive (human-in-the-loop) annotation typically yields better results for these segmentation tasks by providing labels to correct errors in a targeted manner. Simulating such labeling schemes is out-of-scope for our work and would also limit the fair comparison between models. Instead, our evaluation schemes provide an objective evaluation of different methods for pixel (and object) classification that will inform practical applications of these methods.

\FloatBarrier
\subsection{Datasets} \label{app:data}

We evaluate pixel- and object-level classification on five bioimaging datasets spanning a broad range of applications. The datasets were selected to cover diverse acquisition modalities and use cases, including label-free microscopy, fluorescence microscopy, histopathology, and high-content screening. All datasets and their respective classes are listed in table \ref{tab:class_instance_counts}. 

For the CRC~\cite{crc} and HBM~\cite{hbm} datasets, additional class preprocessing was required. Specifically, the original dataset labels were mapped to a reduced set of semantic classes, as detailed in Tables \ref{tab:crc_class_mapping} and \ref{tab:hbm_class_mapping}. Subsequently, classes that were either under-represented or semantically irrelevant, such as labels corresponding to undefined or ambiguous objects, were excluded from the analysis (see Tab.~\ref{tab:class_instance_counts}).

In addition, both CRC and HBM provide multi‑channel images. In this work, we adopted a simple aggregation strategy by summing all channels into a single input channel. Although straightforward, this approach does not exploit channel‑specific information. Future work could investigate more expressive alternatives, such as independent per‑channel processing or clustering channels into a three‑channel (RGB‑like) representation. An overview of the number of image channels and patch sizes for each dataset is given in Table \ref{tab:datasets_format}.

Each dataset was split into fixed training, validation, and test sets (see Table \ref{tab:datasets_splits}), which were kept consistent across all experiments.

% ---------- Table 2: train/val/test split ----------
\begin{table}[h]
\centering
\small
\setlength{\tabcolsep}{6pt}
\renewcommand{\arraystretch}{1.15}
\caption{Train/validation/test splits per dataset (with instance counts).}
\begin{tabular}{@{}p{2.1cm} p{3.5cm} p{3.5cm} l@{}}
\toprule
Dataset & Train split & Val split & Test split \\
\midrule
CRC & 1316 images & 165 images & 166 images \\
& 311,993 inst. & 40,279 inst. & 39,243 inst. \\
HBM & 10910 images & 1634 images & 3030 images \\
& 826,419 inst. & 118,739 inst. & 125,024 inst. \\
PanNuke & 2656 images & 2523 images & 2722 images \\
& 63,218 inst. & 59,872 inst. & 66,654 inst. \\
LIVECell & 3158 images & 569 images & 1512 images \\
& 978,433 inst. & 435,318 inst. & 174,469 inst. \\
Planari & 10460 images & 1307 images & 1309 images \\
& 10460 inst. & 1307 inst. & 1309 inst. \\
\bottomrule
\end{tabular}
\label{tab:datasets_splits}
\end{table}

\begin{table}[h]
\centering
\caption{Number of instances per dataset and class. Classes that are marked with $^\dagger$ are not considered in the classification because they are either underrepresented or irrelevant (e.g. Undetermined)}
\small
\setlength{\tabcolsep}{6pt}
\renewcommand{\arraystretch}{1.15}
\begin{tabular}{@{}p{2.1cm} p{5.6cm} r@{}}
\toprule
Dataset & Class & Number of Instances \\
\midrule

CRC~\cite{crc} & Smooth Muscle Cells & 26,914 \\
 & Macrophages & 41,716 \\
 & Granulocytes & 20,626 \\
 & Plasma Cells & 8,337 \\
 & Neoplastic Cells & 39,184 \\
 & B-Cells & 12,599 \\
 & Nerves$^\dagger$ & 651 \\
 & CD4+ T Cells & 18,159 \\
 & T-Reg & 2,745 \\
 & CD8+ T Cells & 16,333 \\
 & Adipocytes$^\dagger$ & 1,709 \\
 & Others$^\dagger$ & 49,579 \\
\midrule

HBM~\cite{hbm} & HSPC & 6,702 \\
 & Lymphoid & 158,878 \\
 & Myeloid & 417,495 \\
 & Meg/E & 206,656 \\
 & Mesenchymal & 62,761 \\
 & Endothelial & 23,586 \\
 & Muscle & 2,449 \\
 & Neural$^\dagger$ & 52 \\
 & Undetermined$^\dagger$ & 57,535 \\
\midrule

PanNuke~\cite{pannuke}& Neoplastic & 77,403 \\
 & Inflammatory & 32,276 \\
 & Connective/soft tissue & 50,585 \\
 & Epithelial & 26,572 \\
 & Dead cells & 2,908 \\
\midrule

LIVECell~\cite{livecell} & A172 & 130,331 \\
 & BT474 & 128,919 \\
 & BV2 & 351,630 \\
 & Huh7 & 34,107 \\
 & MCF7 & 345,819 \\
 & SHSY5Y & 262,519 \\
 & SkBr3 & 247,779 \\
 & SKOV3 & 86,116 \\
\midrule

Planari & Phenotypically Altered & 12,239 \\
 & Wildtype & 837 \\
\bottomrule
\end{tabular}
\label{tab:class_instance_counts}
\end{table}

\begin{table}[h]
\centering
\small
\setlength{\tabcolsep}{6pt}
\renewcommand{\arraystretch}{1.15}
\caption{Mapping from original CRC annotations to the reduced set of classes.}
\begin{tabular}{@{}p{6.8cm} p{5.8cm}@{}}
\toprule
Original label & Mapped class \\
\midrule
granulocytes & Granulocytes \\
vasculature & Others \\
CD4+ T cells CD45RO+ & CD4+ T cells \\
tumor cells & Neoplastic Cells \\
stroma & Others \\
CD68+CD163+ macrophages & Macrophages \\
adipocytes & Adipocytes \\
plasma cells & Plasma cells \\
CD8+ T cells & CD8+ T cells \\
dirt & Others \\
Tregs & Treg \\
CD4+ T cells & CD4+ T cells \\
CD11c+ DCs & Others \\
B cells & B cells \\
CD11b+CD68+ macrophages & Macrophages \\
smooth muscle & Smooth muscle cells \\
undefined & Others \\
tumor cells / immune cells & Others \\
immune cells / vasculature & Others \\
immune cells & Others \\
NK cells & Others \\
nerves & Nerves \\
CD68+ macrophages GzmB+ & Macrophages \\
CD68+ macrophages & Macrophages \\
lymphatics & Others \\
CD11b+ monocytes & Others \\
CD4+ T cells GATA3+ & CD4+ T cells \\
CD163+ macrophages & Macrophages \\
CD3+ T cells & Others \\
\bottomrule
\end{tabular}
\label{tab:crc_class_mapping}
\end{table}

\begin{table}[h]
\centering
\caption{Mapping from original HBM annotations to the aggregated class set used in this work.}
\small
\setlength{\tabcolsep}{6pt}
\renewcommand{\arraystretch}{1.15}
\label{tab:hbm_class_mapping}
\begin{tabular}{@{}p{6.8cm} p{5.8cm}@{}}
\toprule
Original label & Mapped class \\
\midrule
AEC & Endothelial \\
Adipo-MSC & Mesenchymal \\
Adipocyte & Mesenchymal \\
Artifact & Mesenchymal \\
Autofluorescent & Undetermined \\
B-Cells & Lymphoid \\
CD34+ CD61+ & Meg/E \\
CD4+ T-Cell & Lymphoid \\
CD44+ Undetermined & Undetermined \\
CD8+ T-Cell & Lymphoid \\
CLP & Lymphoid \\
Early Myeloid Progenitor & Myeloid \\
Endosteal & Mesenchymal \\
Erythroblast & Meg/E \\
Erythroid & Meg/E \\
GATA1neg\_Mks & Meg/E \\
GATA1pos\_Mks & Meg/E \\
GMP & Myeloid \\
GMP/Myeloblast & Myeloid \\
HSC & HSPC \\
HSPC & HSPC \\
Immature\_B\_Cell & Lymphoid \\
Intermediate Myeloid & Myeloid \\
MEP/Early Erythroblast & Meg/E \\
Macrophages & Myeloid \\
Mature Myeloid & Myeloid \\
Monocytes & Myeloid \\
Non-Classical Monocyte & Myeloid \\
Plasma Cells & Lymphoid \\
SEC & Endothelial \\
SPINK2+ HSPC & HSPC \\
Schwann Cells & Neural \\
THY1+ MSC & Mesenchymal \\
Undetermined & Undetermined \\
VSMC & Muscle \\
pDC & Myeloid \\
\bottomrule
\end{tabular}
\end{table}

\FloatBarrier

% ---------- Table 3: channels and patch size ----------
\begin{table}[h]
\centering
\small
\setlength{\tabcolsep}{6pt}
\renewcommand{\arraystretch}{1.15}
\caption{Patch size and image channels per dataset.}
\begin{tabular}{@{}p{2.2cm} p{2.2cm} p{4.0cm}@{}}
\toprule
Dataset & Patch size & Channels \\
\midrule
CRC & 512$\times$512 & Multichannel (56 channels) \\
HBM & 512$\times$512 & Multichannel (53 channels) \\
PanNuke & 256$\times$256 & RGB \\
LIVECell & 520$\times$704 & Grayscale \\
Planari & 512$\times$512 & RGB \\
\bottomrule
\end{tabular}
\label{tab:datasets_format}
\end{table}

\FloatBarrier
\subsection{Evaluation Metrics} \label{app:metrics}
We evaluate model performance using the F1 score. In pixel classification, the F1 score is computed at the pixel level by comparing predicted and ground-truth labels across all pixels in the image. In object classification, the F1 score is computed per classified object.

The F1 score is defined as the harmonic mean of precision and recall:
\begin{equation}
    \text{F1} = 2 \cdot \frac{\text{precision} \cdot \text{recall}}{\text{precision} + \text{recall}}
\end{equation}
where, for a given class $ c $,
\begin{align}
    \text{precision}_c &= \frac{TP_c}{TP_c + FP_c}, \\
    \text{recall}_c &= \frac{TP_c}{TP_c + FN_c}.
\end{align}
Here, $ TP_c $, $ FP_c $, and $ FN_c $ denote the number of true positives, false positives, and false negatives for class $ c $, respectively. The final F1 score is computed as the macro-averaged F1 across all classes.

\subsection{Classical Baseline Features}\label{app:classical_feats}

Table \ref{tab:filters} summarizes the hand‑crafted features used for the classical baseline methods. For pixel‑level classification with ilastik, standard multi‑scale image features provided by the Vigra backend were employed, including Gaussian smoothing, Laplacian of Gaussian, gradient‑based features, and second‑order structure descriptors. For object‑level classification, ilastik’s internal feature set was used, comprising geometric, intensity‑based, and shape‑related object descriptors. In addition, object‑level baselines based on scikit‑image’s regionprops were implemented using commonly used morphological and intensity features, such as area, shape descriptors, and intensity statistics.

\begin{table}[h]
\centering
\small
\setlength{\tabcolsep}{6pt}
\renewcommand{\arraystretch}{1.15}
\caption{Overview of the hand‑crafted feature sets used for the classical baseline methods. The table lists the features employed for pixel‑level and object‑level classification using ilastik~\cite{ilastik} and object‑level classification using regionprops from scikit‑image~\cite{skimage}. 
For pixel‑level classification with ilastik, Vigra features were computed at Gaussian scales $\sigma \in \{0.5, 1.0, 2.0, 4.0\}$.}
\begin{tabular}{@{}p{4cm} p{3cm} l@{}}
\toprule
Methods & Implementation & Features \\
\midrule

ilastik & Vigra & Gaussian Smoothing \\
(pixel classification) & & Laplacian of Gaussian \\
& & Gaussian Gradient Magnitude \\
& & Distance of Gaussians \\
& & Structure Tensor Eigenvalues \\
& & Hessian of Gaussian Eigenvalues \\
\midrule
ilastik & ilastik (internal) & Object Area \\
(object classification) & & Mean Intensity \\
& & Length of the Skeleton \\
& & Diameter \\
& & Euclidean Diameter \\
& & Bounding Box Maximum \\
& & Bounding Box Minimum \\
& & Principal components of the object \\
& & Maximum Intensity \\
& & Minimum Intensity \\
& & Center of the object \\
\midrule
RegionProps & scikit-image & area \\
(object classification)& & mean\_intensity \\
& & perimeter \\
& & eccentricity \\
& & solidity \\
& & extent \\
& & major\_axis\_length \\
& & minor\_axis\_length \\
& & orientation \\
& & max\_intensity \\
& & min\_intensity \\
& & centroid \\
\bottomrule
\end{tabular}
\label{tab:filters}
\end{table}

\FloatBarrier
\subsection{Ablation}

\subsubsection{Object Feature Computation}\label{app:feature_ablation}
When performing object-level classification with foundation models, we project instance segmentation masks onto the corresponding embedding space to extract per-object feature representations. For computational efficiency, we aggregate the pixel-level embeddings by computing simple statistical summaries across each object’s region. Table~\ref{tab:object_feature_computation} compares various aggregation strategies on the LIVECell and PanNuke datasets. Results show that using the mean embedding alone yields the best performance on PanNuke, while combining mean and area (object size) leads to optimal results on LIVECell. Given that object size is a biologically meaningful and highly discriminative feature in microscopy, often correlating with cell type, stage, or morphology, we adopt the mean and area aggregation strategy across all experiments.

% from master thesis
\begin{table}[h]
\centering
\caption{Ablation of object‑level feature aggregation strategies on the PanNuke and LIVECell datasets. Object features are constructed by aggregating pixel‑level embeddings using different combinations of statistical descriptors.}\label{tab:object_feature_computation}
\begin{tabular}{l|c|c|c|r|r}
\hline
\textbf{Dataset} & \textbf{Mean} & \textbf{Std} & \textbf{Area} & \textbf{Accuracy} & \textbf{Weighted F1} \\
\hline

 PanNuke & x &   &   & 0.6198 & \textbf{0.6243} \\
         & x & x &   & 0.6024 & 0.607 \\ 
         & x &   & x & 0.5683 & 0.5794 \\ 
         & x & x & x & 0.6029 & 0.6075 \\ \hline 
 LIVECell & x &   &   & 0.8942 & 0.8953 \\
          & x & x &   & 0.8843 & 0.8853 \\ 
          & x &   & x & 0.8952 & \textbf{0.8964} \\  
          & x & x & x & 0.884 & 0.885 \\ \hline

\end{tabular}
\end{table}

\subsubsection{Upsampling with AnyUp}\label{app:anyup_results}

\begin{figure}[h]
  \centering
  \includegraphics[width=\linewidth]{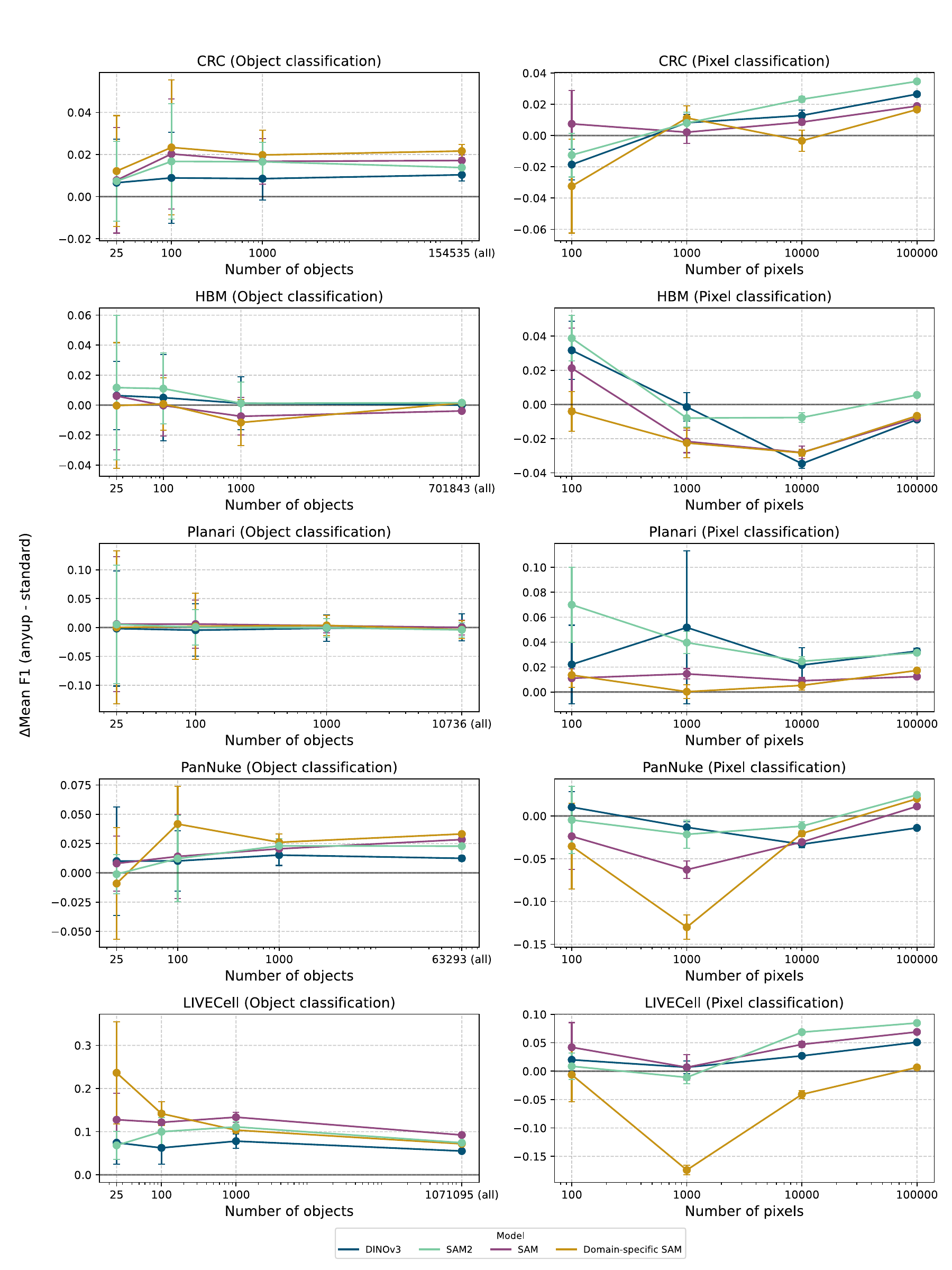}
  \caption{Comparison of standard upsampling via interpolation and AnyUp upsampling for embedding‑based feature extraction. Differences in mean F1 scores across training runs are shown for all datasets. AnyUp provides a noticeable performance gain on LIVECell, with no significant differences observed on the other datasets.}
  \label{fig:anyup}
\end{figure}

We evaluate the effect of AnyUp upsampling (Sec.~\ref{sec:any-up}) for feature extraction in the random forest–based methods. AnyUp is used to upsample the embedding representations prior to feature extraction. The underlying model is pretrained and applied without further training. Figure \ref{fig:anyup} compares the mean F1 scores across all training runs obtained with standard interpolation‑based upsampling and with AnyUp upsampling. A clear performance improvement is observed on the LIVECell dataset when using AnyUp, whereas no significant differences are observed for the remaining datasets. Based on this consistent advantage on LIVECell and the absence of negative effects on other datasets, we employ AnyUp upsampling for all experiments.

\end{document}